\documentclass[sigconf,authorversion]{acmart}
\usepackage{multirow}
\usepackage{bbding}
\usepackage{ulem} 
\usepackage{multicol} 
\usepackage{bbm}
\usepackage{amsmath}
\usepackage{booktabs} 
\usepackage{array} 
\AtBeginDocument{%
  \providecommand\BibTeX{{%
    \normalfont B\kern-0.5em{\scshape i\kern-0.25em b}\kern-0.8em\TeX}}}

\acmConference[MM '20]{MM '20: ACM Multimedia conference}{October 12--16, 2020}{Seattle, United States}

\begin{document}


\title{Dual Gaussian-based Variational Subspace Disentanglement for Visible-Infrared Person Re-Identification}

\author{Nan Pu}
\email{n.pu@liacs.leidenuniv.nl}
\affiliation{%
  \institution{LIACS Media Lab, Leiden University}
}
\author{Wei Chen}
\email{w.chen@liacs.leidenuniv.nl}
\affiliation{%
  \institution{LIACS Media Lab, Leiden University}
}
\author{Yu Liu}
\email{yu.liu@esat.kuleuven.be}
\affiliation{%
  \institution{ESAT-PSI, KU Leuven University}
}

\author{Erwin M. Bakker}
\email{erwin@liacs.leidenuniv.nl}
\affiliation{%
  \institution{LIACS Media Lab, Leiden University}
}
\author{Michael S. Lew}
\email{m.s.k.lew@liacs.leidenuniv.nl}
\affiliation{%
  \institution{LIACS Media Lab, Leiden University}
}

\begin{abstract}
Visible-infrared person re-identification (VI-ReID) is a challenging and essential task in night-time intelligent surveillance systems. Except for the intra-modality variance that RGB-RGB person re-identification mainly overcomes, VI-ReID suffers from additional inter-modality variance caused by the inherent heterogeneous gap. To solve the problem, we present a carefully designed \textit{dual Gaussian-based variational auto-encoder} (DG-VAE), which disentangles an identity-discriminable and an identity-ambiguous cross-modality feature subspace, following a mixture-of-Gaussians (MoG) prior and a standard Gaussian distribution prior, respectively. Disentangling cross-modality identity-discriminable features leads to more robust retrieval for VI-ReID. To achieve efficient optimization like conventional VAE, we theoretically derive two variational inference terms for the MoG prior under the supervised setting, which not only restricts the identity-discriminable subspace so that the model explicitly handles the cross-modality intra-identity variance, but also enables the MoG distribution to avoid posterior collapse. Furthermore, we propose a \textit{triplet swap reconstruction} (TSR) strategy to promote the above disentangling process. Extensive experiments demonstrate that our method outperforms state-of-the-art methods on two VI-ReID datasets.
\vspace{-0.5em}
\end{abstract}

\ccsdesc[500]{Information systems~Information retrieval\vspace{-0.5em}}

\vspace{-1em}
\keywords{cross-modality person re-identification, variational auto-encoder, disentangled representation, Gaussian mixture model}


\maketitle

\vspace{-1em}
\section{INTRODUCTION}
\fancyhead{}
\begin{figure}[ht]
  \centering
  \includegraphics[width=0.48\textwidth]{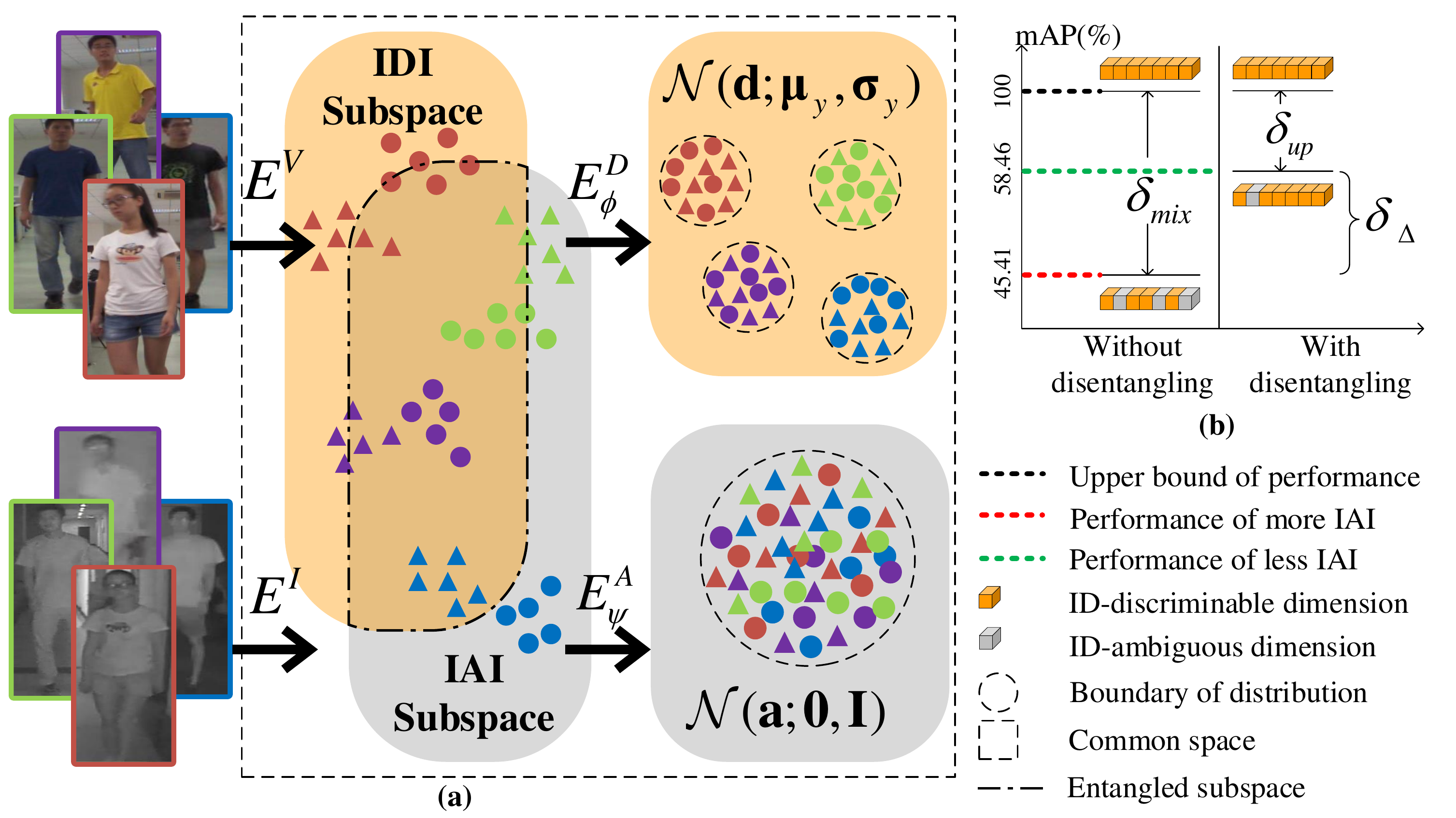}
   \captionsetup{belowskip=-19pt,aboveskip=0pt}
  \caption{\textbf{(a)} Conceptual illustration of the dual Gaussian-based variational subspace disentanglement\label{pic:main1a}. Our DG-VAE factorizes identity-discriminable information (IDI) and identity-ambiguous information (IAI) from the entangled common space of visible-infrared features. The IDI follows the mixture-of-Gaussians prior and the IAI follows the standard Gaussian distribution prior. \textbf{(b)} shows the effectiveness of disentangling methods\label{pic:main1b}. After excluding the IAI from the learned representation, DG-VAE approximates the degraded performance ($\delta_{mix}$) to the IDI-based performance ($\delta_{up}$).}
\end{figure}

\par
Person re-identification (Re-ID), which aims at associating the same pedestrian images across disjoint camera views, has received ever-increasing attention from the computer vision community \cite{zheng2016person,wang2019beyond,leng2019survey}. Most recent efforts \cite{zheng2015scalable,chen2018group,lawen2019attention, zheng2019joint,chen2019mixed,chen2019abd,wu2019unsupervised,huang2019sbsgan,li2019cross,xia2019second} are focused on single-modality image retrieval, \textit{e.g.}, RGB-RGB image matching, which depends on good visible light conditions to catch the appearance of pedestrian. However, visible images might have inferior quality due to inadequate illumination, making it challenging to perform Re-ID by using existing RGB-RGB methods. In practical scenarios, to overcome this limitation, many surveillance cameras can automatically toggle their mode from the visible modality to infrared. This technique enables these cameras to work in dark indoor scenes or at night. Taking advantage of multiple modality cameras, Wu \textit{et~al.} \cite{Wu_2017_ICCV} introduce a challenging RGB-infrared cross-modality person re-identification task, \textit{i.e.}, \textit{visible-infrared person re-identification} (VI-ReID). 
Given an infrared (IR) image of a certain person as a probe image, the goal of VI-ReID is to retrieve the corresponding visible (RGB) image of the same person.

\par
Except for the intra-modality variance involved in single-modality ReID, VI-ReID encounters the additional cross-modality discrepancies resulting from the natural difference between the reflectivity of visible spectrum and the emissivity of thermal spectrum \cite{sarfraz2017deep}. To tackle these two co-existing problems, we approach the VI-ReID task from an information disentanglement perspective. The foundation of associating visible and infrared images relies on semantic information shared across different modalities and the clues given by this information that establish cross-modality connections. We call this kind of information cross-modality \textit{identity-discriminable information} (IDI), \textit{e.g.}, the shape and outline of a body, and some latent characteristics. Unfortunately, there also exists substantial information only belonging to each modality, \textit{e.g.}, color information for visible images and thermal information for infrared images. We call this kind of information the cross-modality \textit{identity-ambiguous information} (IAI). The retrieval performance will be amortized due to noise caused by the redundant IAI dimensions, as illustrated in Fig.~\ref{pic:main1b}. Thus, learned disentanglement representations are supposed to eliminate the IAI and only preserve cross-modality IDI.

\par
Due to the powerful capability of generalization and compaction, variational auto-encoder (VAE) is widely employed to disentanglement representation learning~\cite{liu2018exploring}. However, according to \cite{lucas2019don}, conventional VAEs embed multiple classes or data clusters through a standard Gaussian distribution, which is able to model common characteristics of all inputs and restrict the scope of the distribution. For VI-ReID, the standard Gaussian distribution is effective for IAI, but is relatively ineffective for IDI which often incorrectly handles the structural discontinuity between disparate classes in a latent space. Considering that Mixture-of-Gaussians (MoG) model favorably handles the multi-cluster data, embedding IDI by MoG distribution not only explicitly models the intra-class variations but also ensures the inter-class separability.

\par
To this end, we exploit a dual Gaussian-based VAE (DG-VAE), which aims at disentangling the cross-modality feature maps into IDI and IAI codes for the robust VI-ReID task. Specifically, we enforce the IDI codes follow the MoG distribution where each component corresponds to a particular identity. The variance within each component models the intra-identity differences. Different from the IDI codes, the IAI codes are required to follow the normal distribution. Meanwhile, aligning both modalities with the 
normal distribution is beneficial in regularizing the disentangling process in case reconstruction relies on only the IAI codes. Furthermore, we propose a \textit{triplet swap reconstruction} (TSR) strategy to keep identity-consistency while squeezing IDI and IAI into separate branches, which further promotes disentangling process. To efficiently optimize the proposed DG-VAE, 
we derive a MoG prior term and a maximum entropy regularizer from maximizing the evidence lower bound (ELBO) based on variational bayesian inference. Considering that our DG-VAE is trained in a supervised manner, we introduce a more powerful adaptive large-margin constraint term to substitute the maximum entropy regularizer, which prevents the MoG distribution to occur ``posterior collapse''.

\par
The contributions of this paper are summarized below:
\begin{itemize}
\item We present a novel dual Gaussian-based variational disentanglement architecture with the effective triplet swap reconstruction strategy for addressing the VI-ReID task.
Such design aims at disentangling identity-discriminable and identity-ambiguous feature subspaces to reduce the modality gap.
\item We theoretically derive the variational inference terms for the proposed MoG prior under supervised setting so that our DG-VAE can be efficiently optimized like standard VAE.
\item We experiment with two popular benchmarks where our proposed DG-VAE achieves state-of-the-art performance.
\end{itemize}

\vspace{-1.6em}
\section{RELATED WORK}
\par
\textbf{Visible-infrared person re-identification.} Most existing VI-ReID methods could be divided into two groups. The first group~\cite{Wu_2017_ICCV,ye2018visible,ye2018hierarchical,dai2018cross,hao2019hsme,ye2019modality,hao2019dual} used only feature-level constraints to reduce the intra- and inter-modality discrepancies, which are similar to RGB-RGB Re-ID methods. For example, Wu \textit{et al.} \cite{Wu_2017_ICCV} analyzed three different network structures and used a deep zero padding method for evolving domain-specific structures. Recently, Hao \textit{et al.} proposed a hyper-sphere manifold embedding \cite{hao2019hsme} and dual alignment embedding\cite{hao2019dual} to handle intra-modality and inter-modality variations. The second group~\cite{wang2019learning,wang2019rgb,wang2020cross} incorporated generative adversarial network (GAN) to achieve image-level constraints. For instance, Wang \textit{et~al.} \cite{wang2019rgb} utilized single-direct alignment strategy, in which IR-RGB retrieval is implemented by matching IR images and fake-IR images generated by corresponding RGB images.

\begin{figure}[tb]
  \centering
  \includegraphics[width=0.45\textwidth]{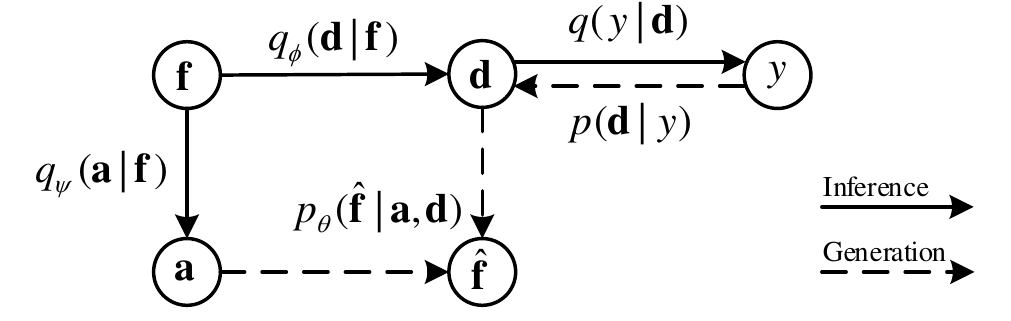}
  \captionsetup{belowskip=-10pt,aboveskip=-2pt}
  \caption{Illustrates the proposed probabilistic graphical model for generative process and inference process.\label{pic:pgm}}
  \vspace{-0.6em}
\end{figure}

\par
The JSIA\cite{wang2020cross} is similar to our DG-VAE that both adopt feature disentanglement technique for VI-ReID but it focuses on factorizing modality-invariant and modality-specific information instead of our identity-discriminable and identity-ambiguous information. Moreover, JSIA employed a image stylization architecture to generate cross-modality images by predicting the parameters of AdaIN\cite{DBLP:conf/iccv/HuangB17}. \textit{In contrast, our DG-VAE adopts a symmetrical encoder-decoder architecture without AdaIN layer to reconstruct feature maps instead of treating the cross-modality variations as only different styles of images. Empirical results demonstrate our method obtains a better performance than JSIA\cite{wang2020cross} as shown in Table~\ref{tab:state}}.

\begin{figure*}[ht]
  \centering
   \captionsetup{belowskip=-12pt,aboveskip=-2pt}
  \includegraphics[width=\textwidth]{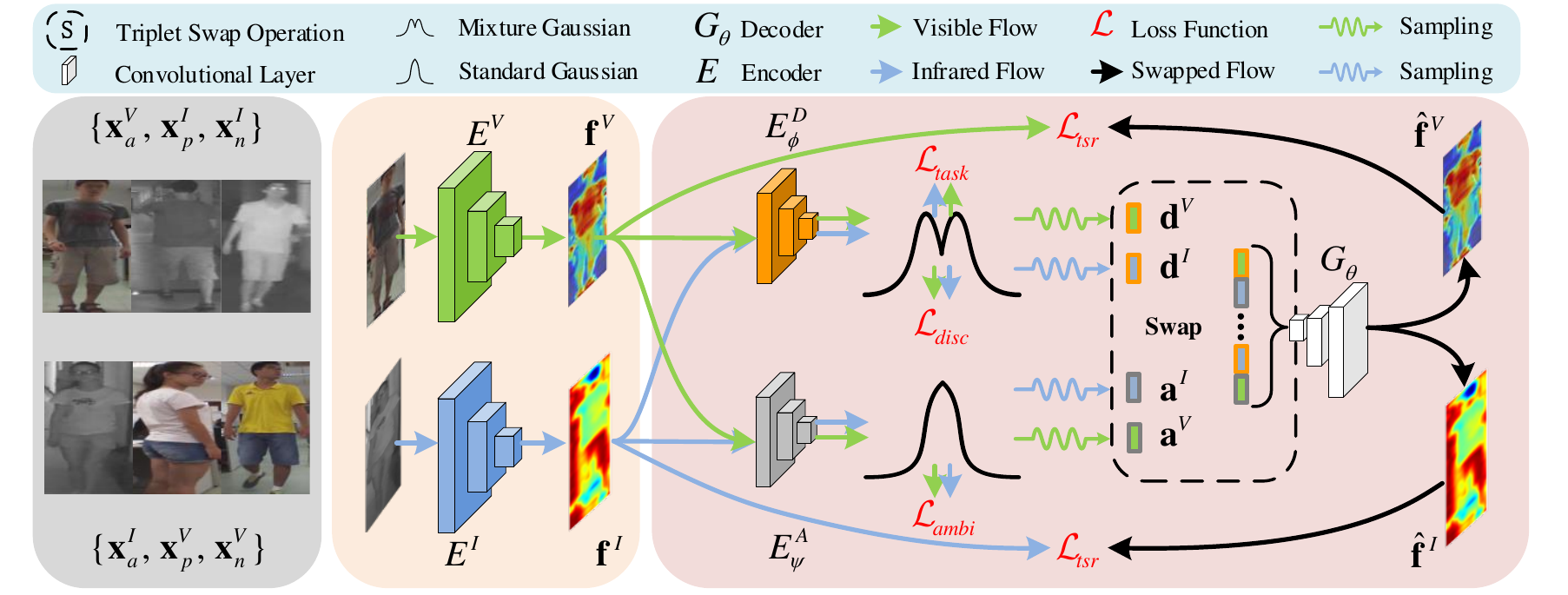}
  \caption{Overview architecture of our proposed DG-VAE\label{pic:2}. Firstly, the cross-modality images triplet are sampled by the strategy illustrated in the left subfigure. We denote the visible image and the infrared image as $\mathbf{X}^{V}_{\kappa}$ and $\mathbf{X}^{I}_{\kappa}$, respectively, where the subscript ${\kappa } \in \{a,p,n\}$ indicates anchor, positive, and negative images. Secondly, the cross-modality triple feature maps ($\mathbf{f^{V}}$ and $\mathbf{f^{I}}$) are extracted by two modality-specific feature extractors, $\mathit{E}^{V}$ and $\mathit{E}^{I}$. Thirdly, the two separated encoders, $\mathit{E}^{D}_{\phi}$ with parameters $\phi$ and $\mathit{E}^{A}_{\psi}$ with parameters $\psi$, encode a feature map to ID-discriminable code ($\mathbf{d^{V}}$ and $\mathbf{d^{I}}$) and ID-ambiguous code ($\mathbf{a}^{V}$ and $\mathbf{a}^{I}$), respectively. Fourthly, the decoder $\mathit{G}_{\theta}$ parametrized by $\theta$, recovers from combining two types of codes to the corresponding reconstructed feature maps ($\hat{\mathbf{f}}^{V}$ and $\hat{\mathbf{f}}^{I}$) by using the proposed TSR strategy.}
\end{figure*}

\par
\textbf{Disentangled representation with VAE.} For learning a disentangled representation, one widely used architecture is VAE \cite{KingmaW13}. Higgins \textit{et.al.} \cite{HigginsMPBGBML17} introduce an adjustable hyperparameter $\beta$ to balance the reconstruction and disentanglement quality and propose an unsupervised VAE-based framework named $\beta$-VAE. 
The drawback of $\beta$-VAE is that a better disentangling result is on the expense of worse feature reconstruction. To solve this problem, Kim \textit{et~al.} \cite{KimM18} introduce a new penalty that provides a better trade-off.
\par
Afterward, the latent 
subspace is disentangled based on the specified and unspecified factors \cite{mathieu2016disentangling}, which aspect is similar to our work. But it employed only an normal distribution prior instead of a MoG prior. Furthermore, our method reconstructs only the modality-specific features instead of the original images, which reduces the amount of parameters and computational cost. We compare two strategies in Sec. \ref{subsec:discussion}. The results in Table~\ref{tab:gan} demonstrate that our method with lightweight decoder still produces the competitive performance compared to image generation methods.

\vspace{-0.5em}
\section{Dual Gaussian-based VAE}\label{sec:DGVAE}
\textbf{Overview.} 
The DG-VAE involves the inference process and generative process (see Fig.~\ref{pic:pgm}) whereby we introduce the dual guassian-based method. First, we adopt the two stream architecture composed of several pre-trained residual blocks as the modality-specific feature extractors, to extract visible-infrared feature maps: $\mathbf{f}^{V}_{\kappa } = \mathit{E}^{V}(\mathbf{X}^{V}_{\kappa })$ and $\mathbf{f}^{I}_{\kappa } = \mathit{E}^{I}(\mathbf{X}^{I}_{\kappa })$, where $\mathbf{f} \in \mathbb{R}^{C \times \lceil H/16\rceil \times \lceil W/16\rceil}$ and $C$ is the number of channels of feature maps. $H$ and $W$ are the height and the width of images $\mathbf{X}$. Second, to disentangle the IDI and IAI subspaces, we propose dual Gaussian-based priors to constrain the latent codes $\mathbf{d}$ and $\mathbf{a}$ generated by IDI encoder $\mathit{E}^{D}_{\phi}$ and IAI encoder $\mathit{E}^{A}_{\psi}$, respectively. This results in the IDI code and IAI code conditionally depending on the extracted feature maps, \textit{i.e.}, $\mathbf{d} \sim p_{\phi}(\mathbf{d}|\mathbf{f})$ and $\mathbf{a} \sim p_{\psi}(\mathbf{a}|\mathbf{f})$. We apply a latent code classifier to fit the condition probability $p(y|\mathbf{d})$, which enables our DG-VAE to infer the correct identity for VI-ReID task. Third, we design a decoder $p_{\theta}(\hat{\mathbf{f}}|\mathbf{d},\mathbf{a})$ with the proposed TSR strategy to promote the disentangling process. To be specific, the reconstructed feature map $\hat{\mathbf{f}}$ is generated by sampling an IAI code $\mathbf{a}$ from $p(\mathbf{a})$ and a corresponding identity $y$, then, an IDI code $\mathbf{d}$ is sampled from the conditional distribution $\mathbf{d}\sim p(\mathbf{d}|y)$. The decoder $p_{\theta}(\hat{\mathbf{f}}|\mathbf{d},\mathbf{a})$ maps the combination of $\mathbf{a}$ and $\mathbf{d}$ to a reconstructed feature map $\hat{\mathbf{f}}$. Note that with $\hat{\mathbf{f}}$, $\mathbf{f}$, $\mathbf{d}$ and $\mathbf{a}$ without superscripts and subscripts we denote the extracted feature maps and latent codes drawn from both modalities, not distinguishing whether it is from an anchor, positive or negative images. The overall pipeline is shown in Fig.~\ref{pic:2}. 

\par
Based on the above reconstruction process, we employ the idea of variational inference to maximize the ELBO so that the data log-likelihood $\hat{\mathbf{f}}$ is also maximized like the conventional VAE\cite{KingmaW13}. By using Jensen's inequality, the log-likelihood in our DG-VAE is:

\vspace{-1.5em}
\begin{equation}\label{eq:elbo}
\begin{split}
\log p(\hat{\mathbf{f}}) &\geq \mathbb{E}_{q_{\phi}(\mathbf{d},y|\mathbf{f}),q_{\psi}(\mathbf{a}|\mathbf{f})} [\log \frac{p_{\theta}(\hat{\mathbf{f}}|\mathbf{d},\mathbf{a})p(\mathbf{d},y)p(\mathbf{a})}{q(y|\mathbf{d})q_{\phi}(\mathbf{d}|\mathbf{f})q_{\psi }(\mathbf{a}|\mathbf{f})} ]\\
& = - D_{KL}(q_{\phi}(\mathbf{d},y|\mathbf{f})||p(\mathbf{d},y))\\
& \quad - D_{KL}(q_{\psi}(\mathbf{a}|\mathbf{f})||p(\mathbf{a}))\\
& \quad + \mathbb{E}_{q_{\phi}(\mathbf{d},y|\mathbf{f}),q_{\psi}(\mathbf{a}|\mathbf{f})} [\log p_{\theta} (\hat{\mathbf{f}}|\mathbf{d},\mathbf{a})].\\
\end{split}
\end{equation}
\vspace{-1.0em}
\par
Hence, in our method, ELBO consists of three terms in Eq. \ref{eq:elbo}. We call the first term as MoG prior regularizer, which matches $q_{\phi}(\mathbf{d},y|\mathbf{f})$ to an identity-specific MoG distribution whose mean and covariance are learned with stochastic gradient variational bayes estimator, and is further introduced in Sec. \ref{subsec:discriminable}. We regard the second term as standard Gaussian prior regularizer. It pushes $q_{\psi}(\mathbf{a}|\mathbf{f})$ to align the prior distribution $p(\mathbf{a})$, which is elaborated in Sec. \ref{subsec:ambiguous}. The third term is negative reconstruction error, which measures whether the latent code $\mathbf{d}$ and $\mathbf{a}$ are informative enough to recover the original feature maps. In our case, we propose triplet swap reconstruction to achieve this goal, which is described in Sec.~\ref{subsec:TSR}. The whole network is optimized by the multi-objective learning scheme in Sec.~\ref{subsec:optimization}. All further theoretical proof and derivations are elaborated in Appendix \ref{app:ELBO}.

\vspace{-1em}
\subsection{Mixture-of-Gaussians Prior for IDI Encoder}\label{subsec:discriminable}

Following the structure of conventional VAEs, we further introduce an identity-discriminable encoder $\mathit{E}^{D}_{\phi}$ upon the above feature extractors ($\mathit{E}^{V}$ and $\mathit{E}^{I}$), to learn IDI representation which enables to identify different persons. Specifically, the IDI codes from visible and infrared images are~$ \mathbf{d}^{V}_{\kappa } = \mathit{E}^{D}_{\phi}(\mathbf{f}^{V}_{\kappa })$ and $\mathbf{d}^{I}_{\kappa } = \mathit{E}^{D}_{\phi}(\mathbf{f}^{I}_{\kappa })$, respectively. These cross-modality IDI codes are organised as 
multiple clusters on a manifold, where the IDI encoder is supposed to tackle the intra-class variation while pushing inter-class distance.


\par
Unfortunately, conventional VAEs often fail to correctly handle the structural discontinuity between disparate classes in a latent space since they use only a standard Gaussian distribution to embed multiple classes or clusters of data~\cite{lucas2019don,razavi2018preventing,he2019lagging}. To solve this problem and further improve the representational capability of IDI codes, $\mathbf{d}$, we expect them to follow the MoG prior distribution with identity-specific mean $\boldsymbol{\mu}_{y}$ and unit variance. For simplicity, we ignore the correlation among different dimensions of $\mathbf{d}$, hence the variance is assumed to be diagonal, and the conditional $p(\mathbf{d}|y)$ is thus equal to:
\begin{equation}\label{eq:pdy}
p(\mathbf{d}|y) = \mathcal{N}(\mathbf{d};\boldsymbol{\mu}_{y}, \mathbf{I}).
\end{equation}
\par
Recall that in Eq.~\ref{eq:elbo}, the KL divergence between $q_{\phi}(\mathbf{d},y|\mathbf{f})$ and $p(\mathbf{d},y)$ is minimized. Since it is difficult to compute the conditional joint probability $q_{\phi}(\mathbf{d},y|\mathbf{f})$, a mean-field distribution is used to estimate $q_{\phi}(\mathbf{d},y|\mathbf{f})$ under unsupervised conditions in \cite{JiangZTTZ17}. In contrast, our DG-VAE is trained in a supervised setting and is supposed to maximize the mutual information between a sample and its label. More clarifications are provided in Appendix~\ref{app:itp}. As a consequence, we assume a conditional dependence between $\mathbf{d}$ and the identity $y$, which results in a variational inference model, $q(\mathbf{d},y|\mathbf{f})$, illustrated in Fig.~\ref{pic:pgm} and formulated as follows:
\begin{equation}\label{eq:pdyf}
q(\mathbf{d},y|\mathbf{f}) = q(y|\mathbf{d})q_{\phi}(\mathbf{d}|\mathbf{f}).
\end{equation}
\par
Then, we define the outputs, $\boldsymbol{\mu}_{\phi(\mathbf{f})}$ and $\boldsymbol{\sigma}_{\phi(\mathbf{f})}$, of $\textit{E}^{D}_{\phi}$ as follows:
\begin{equation}\label{eq:pdf}
q_{\phi}(\mathbf{d}|\mathbf{f})= \mathcal{N}(\mathbf{d};\boldsymbol{\mu}_{\phi(\mathbf{f})}, \boldsymbol{\sigma}_{\phi(\mathbf{f})}^{2}).
\end{equation}
\par 
Using Eq.~\ref{eq:elbo}, \ref{eq:pdy}, \ref{eq:pdyf} and \ref{eq:pdf}, we have:
\begin{equation}\label{eq:twotermsKL}
\begin{split}
\quad D_{KL}(q_{\phi}(\mathbf{d},y|\mathbf{f})||p(\mathbf{d},y))=&\sum_{y}q(y|\mathbf{d})D_{KL}(q_{\phi}(\mathbf{d}|\mathbf{f})||p(\mathbf{d}|y))\\
& + D_{KL}(q(y|\mathbf{d})||p(y)),\\
\end{split}
\end{equation}
where the conditional probability $q(y|\mathbf{d})$ in a supervised setting is the one-hot encoding of the label $y$. Thus, the first term is to align the latent distribution $q_{\phi}(\mathbf{d}|\mathbf{f})$ and the corresponding class-specific distribution $p(\mathbf{d}|y)$. We call it as MoG prior term. The second term encourages posterior probability $q(y|\mathbf{d})$ to approximate the uniform distribution $p(y)$. In fact, it corresponds to maximize conditional entropy, when the components of the Gaussian are expected to be separable without overlap, which acts as a maximum entropy regularizer. We provide the proof and derivation of the two terms in Appendix~\ref{app:gmm} and Appendix~\ref{app:mr}, respectively. If these two terms are exploited in an unsupervised manner, they indeed implement a large-margin clustering algorithm.
\par
Considering that our DG-VAE is a kind of supervised model, the maximum entropy regularizer could be more effectively accomplished by using supervision. Thus, we introduce an adaptive large-margin constrain (ALMC) $\mathcal{L}_{lmc}$ to disperse the components of the MoG distribution with the corresponding learnable margins, inspired by~\cite{liu2019adaptiveface}. Finally, by using the reparametrization trick \cite{KingmaW13} and respectively sampling $\mathbf{d}$ and $\mathbf{a}$ from $q_{\phi}(\mathbf{d}|\mathbf{f})$ and $q_{\psi }(\mathbf{a}|\mathbf{f})$, $\mathcal{L}_{gmm}$ and $\mathcal{L}_{lmc}$ can be calculated by the following closed-form solution:
\begin{equation}
\begin{split}
\mathcal{L}_{gmm}=\mathbb{E}_{\mathbf{d}\sim q_{\phi}(\mathbf{d}|\mathbf{f})}\sum_{c}^{N_{y}}\mathbbm{1}(y=c)\Big(\ln \big(det(\boldsymbol{\sigma}_{\phi(\mathbf{f})})\big) +\frac{1}{2} \left \| \mathbf{d} - \boldsymbol{\mu}_{y}\right \| ^{2}\Big),\\
\end{split}
\end{equation}
\begin{equation}\label{eq:result:gmprt}
\mathcal{L}_{lmc} = - \mathbb{E}_{\mathbf{d}\sim q_{\phi}(\mathbf{d}|\mathbf{f})}\log \dfrac{\exp^{-\mathit{D}_{M}(\mathbf{d},y,\mathbf{f})}}{\sum_{c=1,c\neq y}^{N_{y}}\exp^{-\left \|\mathbf{d}-\boldsymbol{\mu}_{c}\right\|^{2}} + \exp^{-\mathit{D}_{M}(\mathbf{d},y,\mathbf{f})}},
\end{equation}
where $N_{y}$ is the number of identities in training set. The indicator function $\mathbbm{1}()$ equals $1$ if $y$ equals $c$; and $0$ otherwise. The identity-specific means $\boldsymbol{\mu}_{y}$ are initialized randomly and updated with gradient descent. Furthermore, given some samples following Mixture-of-Gaussians (MoG) distribution, the samples lie on the MoG manifold. The distance between a re-sampled $\mathbf{d}$ with identity $y$ and the corresponding mean $\boldsymbol{\mu}_{\phi(\mathbf{f})}$ should be the squared Mahalanobis distance:
\begin{equation}\label{eq:margin}
\mathit{D}_{M}(\mathbf{d},y,\mathbf{f}) = (\mathbf{d} - \boldsymbol{\mu}_{\phi(\mathbf{f})})^{T}\boldsymbol{\sigma}_{\phi(\mathbf{f})}^{-1}(\mathbf{d} - \boldsymbol{\mu}_{\phi(\mathbf{f})}) - \alpha_{y},\\
\end{equation}
where the identity-specific margin $\alpha\geq 0$ is a learnable parameter, which can easily achieve the adaptive large-margin MoG constraint, as illuminated in Appendix~\ref{app:tsne}.
\par
The combination of $\mathcal{L}_{gmm}$ and $\mathcal{L}_{lmc}$ not only strengthens the network capability of handling multi-class data, but it also constrains the scope of the distribution of each identity by aligning the prior distributions while pushing the inter-class distances. It is robust to outliers or noise so that it has stronger generalization. The quantitative analysis is elaborated in Sec.~\ref{subsec:discussion}. The objective function of IDI branch is weighted by $\lambda_{gmm}$ and $\lambda_{lmc}$:
\begin{equation}\label{eq:disc}
\mathcal{L}_{disc} = \lambda_{gmm}\mathcal{L}_{gmm} +\lambda_{lmc}\mathcal{L}_{lmc}.
\end{equation}

\begin{figure*}[ht]
  \centering
  \includegraphics[width=\textwidth]{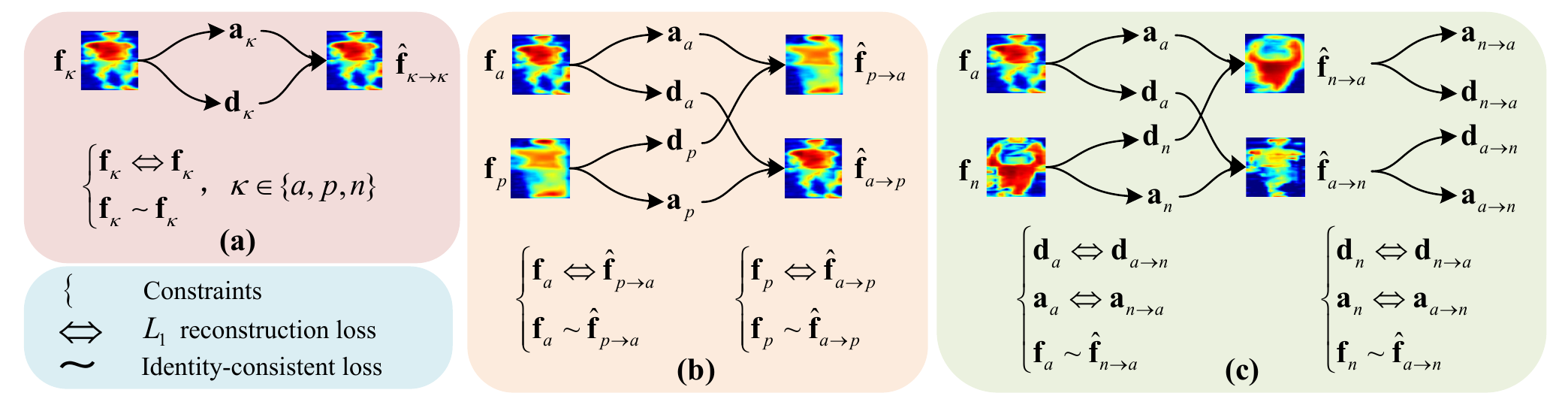}
  \captionsetup{belowskip=-12pt,aboveskip=-4pt}
  \caption{Illustration of the proposed triplet swap reconstruction in detail. For clear representation, we show only the case of the triplet that consists of a visible anchor, infrared positive, and infrared negative. Likewise, the other triplet that contains a infrared anchor, visible positive, and visible negative is swapped and reconstructed.\label{pic:3}}
\end{figure*}

\vspace{-1.0em}
\subsection{Standard Gaussian Prior for IAI Encoder}\label{subsec:ambiguous}
To learn the identity-ambiguous representation, we design an IAI encoder $\mathit{E}^{A}_{\psi}$ to encode the extracted feature maps ($\mathbf{f}^{V}_{\kappa }$ and $\mathbf{f}^{I}_{\kappa })$ to IAI codes, which is fomulated as:~$\mathbf{a}^{V}_{\kappa } = \mathit{E}^{A}_{\psi}(\mathbf{f}^{V}_{\kappa })$ and $\mathbf{a}^{I}_{\kappa } = \mathit{E}^{A}_{\psi}(\mathbf{f}^{I}_{\kappa })$.
\par
Unlike the IDI codes that should provide a more reasonable embedding subspace for discriminating different identities, the IAI representation aims at modeling the common characteristics across all the input images. 
It is scarcely able to identify different persons from these characteristics but they are necessary for feature reconstruction. To this end, we encourage the identity-ambiguous codes to approximate the prior normal distribution $p(\mathbf{a})$ with zero mean and unit variance instead of the MoG prior distribution. We follow \cite{lu2019unsupervised} to define a Gaussian distribution with mean $\boldsymbol{\mu}_{\psi(\mathbf{f})}$ and diagonal covariance $\boldsymbol{\sigma}_{\psi(\mathbf{f})}$, \textit{i.e.}, $q_{\psi}(\mathbf{a}|\mathbf{f})$, as the output of encoder $\mathit{E}^{A}_{\psi}$ parametrized by $\psi$. According to Eq. \ref{eq:elbo}, the ID-ambiguous KL regularization term is formulated as follows,
\begin{equation}\label{eq:kl}
D_{KL} \Big( \mathcal{N} \big( \boldsymbol{\mu}_{\psi(\mathbf{f})},\boldsymbol{\sigma}_{\psi(\mathbf{f})}^{2} \big)\Big|\Big|\mathcal{N} \big( \mathbf{0},\mathbf{I} \big) \Big).
\end{equation}
\par
The KL divergence regularizes the identity-ambiguous codes by limiting the distribution range, such that they do not contain much identity-discriminable information so as to facilitate the disentanglement process \cite{lee2018diverse}. Then, with a reparametrization trick \cite{KingmaW13}, Eq. \ref{eq:kl} can be calculated by the following closed solution:
\begin{equation}\label{eq:kl_rp}
\mathcal{L}_{ambi} =\dfrac{1}{N^{V}+N^{I}}\sum_{i=1}^{N^{V}+N^{I}}\sum_{l=1}^{L} \Big( (\boldsymbol{\mu}_{il})^{2}+(\boldsymbol{\sigma}_{il})^{2}-\log \big( \boldsymbol{\sigma}_{il}\big)^{2}-1  \Big ),\\
\end{equation}
where $N^{V}$ and $N^{I}$ denote the number of visible and infrared images in a mini-batch, respectively. $L$ is the length (dimension) of the identity-ambiguous code. Eq. \ref{eq:kl_rp} allows the latent variational representation to be differentiated and capable of back-propagation.
\par
Furthermore, although we train the encoders separately to extract these information, the decoder $G_{\theta}$ may largely rely on the identity-ambiguous codes to synthesize new feature maps, while ignoring the identity-discriminable ones. This situation will distract the feature disentanglement process during training. Thus, using standard Guassian prior to restrict the scope of feature distribution also enables to effectively mitigate this problem.



\vspace{-0.5em}
\subsection{Triplet Swap Reconstruction}\label{subsec:TSR}
Moreover, we propose TSR strategy to perform the whole encoder-decoder process. Different from conventional VAEs, DG-VAE carries out the reconstruction where the decoder uses two types of latent codes swapped across cross-modality triplet as input.
\par
To squeeze the IDI and IAI contained in visible-infrared feature maps into separate branches, we assume that the disentangled representation satisfies the following conditions: 1) An original feature map should be reconstructed from its ID-discriminable and ID-ambiguous code as illustrated in Fig.~\ref{pic:3}\textbf{(a)}; 2) Swapping ID-discriminable codes between cross-modality anchor and positive feature maps, the positive feature maps should be reconstructed from the swapped ID-discriminable code and their ID-ambiguous code, while the identities of reconstructed feature maps should be kept consistently, as shown in Fig.~\ref{pic:3}\textbf{(b)}; 3) Swapping ID-ambiguous codes between a cross-modality pair (whatever positive pair or negative pair), the identities of reconstructed feature maps should correspond to 
that of the ID-discriminable codes like Fig.~\ref{pic:3}\textbf{(c)}.

\par
To reach these conditions, the decoder $\mathit{G}_{\theta}$ is required to reconstruct an anchor feature map $\mathbf{f}^{V}_{a}$ (or $\mathbf{f}^{I}_{a}$) from $\mathbf{d}^{V}_{a}\odot \mathbf{a}^{V}_{a}$ and $\mathbf{d}^{I}_{p}\odot \mathbf{a}^{V}_{a}$ (or $\mathbf{d}^{I}_{a}\odot \mathbf{a}^{I}_{a}$ and $\mathbf{d}^{V}_{p}\odot \mathbf{a}^{I}_{a}$), where the $\odot $ denotes a concatenation operation as depicted in Fig.~\ref{pic:2}. Thus, the reconstruction loss is:
\begin{equation}\label{eq:rec}
\begin{split}
\mathcal{L}_{rec} &=  \mathbb{E}_{{\kappa }\in \{a,p,n\}}\bigg[ \left \| \mathbf{f}_{\kappa}^{V}-\mathit{G}_{\theta}(\mathbf{d}^{V}_{\kappa}\odot \mathbf{a}^{V}_{\kappa}) \right \|_{1} + \left \| \mathbf{f}_{\kappa}^{I}-\mathit{G}_{\theta}(\mathbf{d}^{I}_{\kappa}\odot \mathbf{a}^{I}_{\kappa}) \right \|_{1}\bigg]\\
& \quad +\mathbb{E}_{\substack{i,j \in \{ a,p\} \\ 
  i\neq j}}\bigg[ \left \| \mathbf{f}_{i}^{V}-\mathit{G}_{\theta}(\mathbf{d}^{I}_{j}\odot \mathbf{a}^{V}_{i}) \right \|_{1} + \left \| \mathbf{f}_{i}^{I}-\mathit{G}_{\theta}(\mathbf{d}^{V}_{j}\odot \mathbf{a}^{I}_{i}) \right \|_{1}\bigg].\\
\end{split}
\end{equation}
\par
The first term performs a standard self-reconstruction, enforcing the combination of ID-discriminable and ID-ambiguous codes from the same image to contain all information to reconstruct the original features. It is employed as the reconstruction error term in Eq.~\ref{eq:elbo}. The second term encourages the encoder $\mathit{E}^{D}_{\phi}$ to extract the shared identity-discriminable features, $\mathbf{d}^{V}_{a}$(or $\mathbf{d}^{I}_{a}$) and $\mathbf{d}^{V}_{p}$(or $\mathbf{d}^{I}_{p}$) from a pair of $\mathbf{f}^{I}_{a}$ (or $\mathbf{f}^{V}_{a}$)  and $\mathbf{f}^{V}_{p}$ (or $\mathbf{f}^{I}_{p}$) , focusing on the consistent information between them. Other factors, not shared by $\mathbf{f}^{V}_{a}$ and $\mathbf{f}^{I}_{a}$, are encoded into the identity-ambiguous features, $\mathbf{d}^{V}_{a}$ (or $\mathbf{d}^{I}_{a}$) and $\mathbf{d}^{I}_{p}$ (or $\mathbf{d}^{V}_{p}$). Note that it might be unreasonable to impose an $L_{1}$ reconstruction constraint on the reconstructed feature maps generated from cross-modality swapped negative pair, \textit{i.e.}, $\mathit{G}_{\theta}(\mathbf{d}^{I}_{a}\odot \mathbf{a}^{V}_{n})$, $\mathit{G}_{\theta}(\mathbf{d}^{I}_{n}\odot \mathbf{a}^{V}_{a})$, $\mathit{G}_{\theta}(\mathbf{d}^{V}_{a}\odot \mathbf{a}^{I}_{n})$, and $\mathit{G}_{\theta}(\mathbf{d}^{V}_{n}\odot \mathbf{a}^{I}_{a})$. Because both their IDI and IAI are not shared, hence, the reconstructed feature maps are hardly recovered from their combinations. 
\par
Furthermore, we introduce the cycle-consistency loss \cite{zhu2017unpaired} to make the reconstructed features preserve the ID-discriminable and ID-ambiguous codes of their original features. Moreover, given the reconstructed feature $\hat{\mathbf{f}}_{\kappa }$ ($\hat{\mathbf{f}}^{V}_{\kappa }$ or $\hat{\mathbf{f}}^{I}_{\kappa }$), the latent code classifier $\mathit{D}_{C}$ enforces the identity-consistent constraints on all of the codes. The cycle- and identity-consistency loss are defined as follows,
\begin{equation}
\begin{split}
\mathcal{L}_{cyc}&= \mathbb{E}_{i,j\in \{a,p,n\}} \bigg[\left \| \mathbf{a}_{i} - \mathit{E}^{A}_{\psi}\big(\mathit{G}_{\theta}(\mathbf{d}_{j}\odot \mathbf{a}_{i})\big) \right \|_{1} \\
& \quad +  \left \| \mathbf{d}_{i } - \mathit{E}^{D}_{\phi}\big(\mathit{G}_{\theta}(\mathbf{d}_{i}\odot \mathbf{a}_{j})\big) \right \|_{1} \bigg],\\
\end{split}
\end{equation}
\begin{equation}
\begin{split}
\mathcal{L}_{idc} =  \mathbb{E}_{{\kappa }\in \{a,p,n\}}\Big[-\log \mathit{D}^{C}\big(\mathit{E}^{D}_{\phi}(\hat{\mathbf{f}}_{\kappa })\big)\Big],\\
\end{split}
\end{equation}
where $\mathit{D}^{C}\big(\mathit{E}^{D}_{\phi}(\hat{\mathbf{f}}_{\kappa })\big)$ is the predicted probability of the current feature map $\hat{\mathbf{f}}_{\kappa }$ belonging to the ground-truth class (identity).
\par
The overall objective for the proposed TSR is given as follows,
\begin{equation}
\mathcal{L}_{tsr} =  \lambda_{idc}\mathcal{L}_{idc} + \lambda_{cyc}\mathcal{L}_{cyc} + \lambda_{rec}\mathcal{L}_{rec},
\end{equation}
where $\lambda_{idc}$, $\lambda_{cyc}$ and $\lambda_{rec}$ are the trade-off factors.


\vspace{-0.5em}
\subsection{Multi-objective Learning and Optimization}\label{subsec:optimization}
\par 
To simultaneously consider the reparameterized latent variate, DG-VAE shares the same feature classifier for both $\boldsymbol{\mu_{\phi(\mathbf{f})}}$ and $\mathbf{d}$, which are optimized by the weighted sum of standard cross-entropy loss and cross-modality triplet loss \cite{dai2018cross}, \textit{e.g.}, $\ell({a}^{V},{p}^{I},{n}^{I})=[D({a}^{V},{p}^{I})- D({a}^{V},{n}^{I})+ m]_{+}$ and $\ell({a}^{I},{p}^{V},{n}^{V})=[D({a}^{I},{p}^{V})- D({a}^{I},{n}^{V})+ m]_{+}$, where $D(\cdot, \cdot)$ denotes $L_{2}$ distance between two normalized inputs and $m$ is a fixed margin hypeparameter. Given a mini-batch with $N^{V}$ visible images and $N^{I}$ infrared images, the retrieval features are formulated as follows,
\begin{equation}
\mathcal{L}_{id}= -\frac{1}{N^{V}+N^{I}}\sum_{i}^{N^{V}+N^{I}}\Big(\log \mathit{D}^{C}(\boldsymbol{\mu_{\phi(\mathbf{f}_{i})}})+\lambda_{aux}\log \mathit{D}^{C}(\mathbf{d}_{i})\Big)\\
\end{equation}
\begin{equation}
\begin{split}
\mathcal{L}_{cmtl} &= \frac{1}{N^{V}}\sum^{N^{V}}_{(a,p,n)}\Big(\ell(\boldsymbol{\mu_{\phi(\mathbf{f}_{a}^{V})}},\boldsymbol{\mu_{\phi(\mathbf{f}_{p}^{I})}},\boldsymbol{\mu_{\phi(\mathbf{f}_{n}^{I})}}) + \lambda_{aux}\ell(\mathbf{d}_{a}^{V},\mathbf{d}_{p}^{I},\mathbf{d}_{n}^{I})\big) \\
& + \frac{1}{N^{I}}\sum^{N^{I}}_{(a,p,n)}\Big(\ell(\boldsymbol{\mu_{\phi(\mathbf{f}_{a}^{I})}},\boldsymbol{\mu_{\phi(\mathbf{f}_{p}^{V})}},\boldsymbol{\mu_{\phi(\mathbf{f}_{n}^{V})}}) + \lambda_{aux}\ell(\mathbf{d}_{a}^{I},\mathbf{d}_{p}^{V},\mathbf{d}_{n}^{V})\Big) \\
\end{split}
\end{equation}
\begin{equation}
\mathcal{L}_{task} = \lambda_{id}\mathcal{L}_{id} + \lambda_{cmtl}\mathcal{L}_{cmtl},
\end{equation}
where $\mathit{D}^{C}(\cdot)$ is the predicted probability of the input belonging to the ground-truth class. $\lambda_{aux}$, $\lambda_{id}$ and $\lambda_{cmtl}$ are trade-off factors.
\par
\textbf{The overall objective} is a weighted sum of all loss functions:
\begin{equation}
\mathcal{L}_{all} = \mathcal{L}_{task} + \lambda_{tsr}\mathcal{L}_{tsr} +  \lambda_{ambi}\mathcal{L}_{ambi} + \lambda_{disc}\mathcal{L}_{disc}
\end{equation}
where $\lambda_{tsr},\lambda_{ambi}$ and $\lambda_{disc}$ are the loss weights to balance each term during the training process detailed in Sec.~\ref{exp:hyper}. We only use the mean of IDI code, $\boldsymbol{\mu_{\phi(\mathbf{f})}}$, as retrieval feature at testing time.

\vspace{-0.5 em}
\section{EXPERIMENTS}

\subsection{Experimental Setup}
\textbf{Datasets and Evaluation Settings.} We evaluated our method on two publicly available datasets: RegDB \cite{nguyen2017person} and SYSU-MM01 \cite{Wu_2017_ICCV}. Our experiments followed the RegDB evaluation protocol as described in \cite{ye2018hierarchical,ye2018visible} and the SYSU-MM01 evaluation protocol from \cite{Wu_2017_ICCV}. The RegDB dataset 
consists of 2,060 visible images and 2,060 far-infrared images with 206 identities for training. The testing set contains 206 identities with 2,060 visible images for the query and 2,060 far-infrared images for the gallery. 
The SYSU dataset contains 22,258 visible images and 11,909 near-infrared images of 395 identities for training. The testing set includes 96 identities with 3,803 near-infrared images for the query and 301 visible images as the gallery. The SYSU dataset is collected by four visible cameras and two near-infrared cameras, used in both indoor and outdoor environments. We adopted the most challenging \textit{single-shot all-search} mode and repeated the above evaluation of 10 trials with a random split of the gallery and probe set to get final results.
\par
\textbf{Evaluation metrics.} We adopt two popular evaluation metrics: rank-k and mean Average Precision (mAP).
The rank-k (dubbed R-1 or R-10) indicates the cumulative rate of true matches in the top-k position. The mAP considers person re-identification as a retrieval task, which reflects the comprehensive retrieval performance.



\vspace{-1.0 em}
\subsection{Implementation Details}
DG-VAE is implemented using the Pytorch framework on an NVIDIA Titan Xp GPU. The source codes will be publicly available at $https://github.com/TPCD/DG-VAE$.
\par
\textbf{Mini-batch organization.} We follow \cite{hao2019dual} and resize visible and infrared images to $3\times 384\times 128$. Each mini-batch contains 4 different identities, while each identity has 4 pairs of visible and infrared images. Within the mini-batch, we form 32 cross-modality triplets by selecting positive and negative instances for each image, following the rule that the anchor and positive are from the same person in different modalities, while anchor and negative are required to have different identification labels in different modalities.
\par
\textbf{Network architecture.} Since the low-level visual patterns (\textit{e.g.}, texture, contour) of infrared images are similar to general visible images \cite{ye2018visible}, the two modality-specific feature extractors share the same architecture, which consists of three pre-trained residual blocks in \cite{wang2018learning} with $C=1024$ channels of output. Note that the parameters of two streams are optimized separately to capture the information of each modality. Moreover, the IDI and IAI encoders respectively contain two pre-trained blocks of ResNet-50\cite{he2016deep} followed by two heads for predicting mean and variance. Both heads of IDI encoder is followed by two max-pooling layer, since it favourably captures the discriminative features. The heads of IAI encoder are followed by two avg-pooling layers, since it could provide a comprehensive representation. Meanwhile, the different-modality feature maps share the same IDI and IAI encoder. The main reason is that both modalities include two such types of information and sharing parameters reduces the computational costs. Finally, the decoder consists of a fully connected layer with batch normalization \cite{ioffe2015batch}, Leaky ReLU \cite{maas2013rectifier}, Dropout \cite{srivastava2014dropout} and a series of transposed convolutional layers. It inputs IDI and IAI features whose dimensions are $2048$ and $512$, respectively. The latent code classifier has only one fully connected layer. The detailed configuration of the architecture is given in Appendix \ref{append:nna}.

\par
\textbf{Training strategy.} We use the Adam optimizer \cite{kingma2014adam} with $\beta_{1} = 0.9$ and $\beta_{2} = 0.999$ to train two extractors, two encoders and a decoder, and use stochastic gradient descent with momentum $0.9$ for the latent code classifier. Similar to the training scheme in \cite{ge2018fd}, we train DG-VAE in three stages. In the first stage, we train the IDI encoder $\mathit{E}_{\phi}^{D}$ using $\mathcal{L}_{task}$ and $\mathcal{L}_{disc}$, for $100$ epochs. A learning rate is set to $2e^{-4}$. In the second stage, we fix the IDI branch, and train the IAI encoder $\mathit{E}_{\psi}^{A}$, the decoder $G_{\theta}$, and the latent code classifier $D^{C}$ with the corresponding losses, $\mathcal{L}_{tsr}$ and $\mathcal{L}_{ambi}$. This process iterates for $50$ epochs with learning rate of $2e^{-4}$. Finally, we train the whole network end-to-end with the learning rate of $2e^{-5}$ for $100$ epochs. We augment the datasets with horizontal flipping and random erasing \cite{zhong2017random}.

\begin{table}[t]
\centering
  \captionsetup{belowskip=-15pt,aboveskip=2pt}
\caption{Comparison with state-of-the-art VI-ReID methods on the RegDB and the SYSU-MM01 datasets. \label{tab:state}}
\setlength{\tabcolsep}{1.70mm}

\begin{tabular}{p{0.95cm}<{\centering}p{1.55cm}p{0.48cm}<{\centering}p{0.59cm}<{\centering}p{0.5cm}<{\centering}p{0.48cm}<{\centering}p{0.59cm}<{\centering}p{0.5cm}<{\centering}}
\toprule[1pt]
\multicolumn{1}{l}{} & Datasets & \multicolumn{3}{c}{RegDB} & \multicolumn{3}{c}{SYSU-MM01} \\  \cmidrule(r){2-2} \cmidrule(r){3-5} \cmidrule(r){6-8}
\multicolumn{1}{l}{} & Methods & R-1 & R-10 & mAP & R-1 & R-10 & mAP \\  \midrule[1pt]
\multirow{2}{*}{ \shortstack{\small $\!\!\!\!\!\!\!$  Feature \\ \small $\!\!\!\!\!\!\!$ Extraction}} & ZERO\citep{Wu_2017_ICCV} & 17.75 & 34.21 & 18.90 & 14.80 & 54.12 & 15.95 \\
 & TONE\cite{ye2018hierarchical} & 16.87 & 34.03 & 14.92 & 12.52 & 50.72 & 14.42 \\ \cmidrule(r){1-8}
\multirow{5}{*}{ \shortstack{\small $\!\!\!\!\!\!\!$  Metric \\ \small $\!\!\!\!\!\!\!$ Learning}} & HCML\cite{ye2018hierarchical} & 24.44 & 47.53 & 20.80 & 14.32 & 53.16 & 16.16 \\
 & BCTR\cite{ye2018visible} & 32.67 & 57.64 & 30.99 & 16.12 & 54.90 & 19.15 \\
 & BDTR\cite{ye2018visible} & 33.47 & 58.42 & 31.83 & 17.01 & 55.43 & 19.66 \\
 & HSME\cite{hao2019hsme} & 41.34 & 65.21 & 38.82 & 18.03 & 58.31 & 19.98 \\
 & D-HSME\cite{hao2019hsme} & 50.85 & 73.36 & 47.00 & 20.68 & 62.74 & 23.12 \\  \cmidrule(r){1-8}
 \multirow{3}{*}{ \shortstack{\small $\!\!\!\!\!\!\!$  Image \\ \small $\!\!\!\!\!\!\!$ Generation}} & D2RL\cite{wang2019learning} & 43.4 & 66.10 & 44.10 & 28.90 & 70.60 & 29.20 \\
 & JSIA\cite{wang2020cross} & 48.10 & - & 48.90 & 38.10 & 80.70 & 36.90 \\ 
 & AlignGAN\cite{wang2019rgb} & 56.30 & - & 53.40 & 42.40 & 85.00 & 40.70 \\
 \cmidrule(r){1-8}
\multirow{3}{*}{ \shortstack{\small $\!\!\!\!\!\!\!\!\!$  Distribution \\ \small $\!\!\!\!\!\!\!$ Alignment}} & cmGAN\cite{dai2018cross} & - & - & - & 26.97 & 67.51 & 27.80 \\ 
 & MAC\cite{ye2019modality} & 36.43 & 62.36 & 37.03 & 33.37 & 82.49 & 44.95 \\
 & DFE\cite{hao2019dual} & 70.13 & 86.32 & 69.14 & 48.71 & 88.86 & 48.59 \\  \cmidrule(r){1-8}
 & DG-VAE & \textbf{72.97} & \textbf{86.89} & \textbf{71.78} & \textbf{59.49} & \textbf{93.77} & \textbf{58.46} \\ \bottomrule[1pt]
\end{tabular}
\vspace{-0.5cm}
\end{table}

\begin{table*}[th]
\center
  \captionsetup{belowskip=-15pt,aboveskip=2pt}
\caption{\label{tab:ablation} Evaluation of our baseline and its variants with different loss functions on the SYSU-MM01 and RegDB dataset.}
\begin{tabular}{cp{0.6cm}<{\centering}p{0.7cm}<{\centering}p{0.6cm}<{\centering}p{0.6cm}<{\centering}p{0.6cm}<{\centering}p{0.7cm}<{\centering}p{0.6cm}<{\centering}p{0.6cm}<{\centering}p{1cm}<{\centering}p{0.6cm}<{\centering}p{1cm}<{\centering}p{0.6cm}<{\centering}}
\toprule[1pt]
& \multicolumn{8}{c}{Loss Functions} & \multicolumn{2}{c}{SYSU-MM01} & \multicolumn{2}{c}{RegDB} \\  \cmidrule(r){2-9} \cmidrule(r){10-11} \cmidrule(r){12-13}
& $\mathcal{L}_{id}$&$\mathcal{L}_{cmtl}$&$\mathcal{L}_{rec}$& $\mathcal{L}_{idc}$ & $\mathcal{L}_{cyc}$&$\mathcal{L}_{ambi}$ &$\mathcal{L}_{gmm}$&$\mathcal{L}_{lmc}$& Rank-1 & mAP & Rank-1 & mAP \\ \midrule[1pt]
\shortstack{Baseline}  &\Checkmark &\Checkmark & & & & & & &44.45&44.89&58.74&59.80 \\ \cmidrule(r){1-13}
\multirow{2}{*}{\shortstack{\small{Mixture-of-Gaussians Prior}}}  & \Checkmark &\Checkmark & & & & &\Checkmark & &42.12&42.37&58.12&59.24\\ 
 & \Checkmark &\Checkmark & & & & &\Checkmark &\Checkmark &45.39&45.41&60.28&60.92\\ \cmidrule(r){1-13}
 \multirow{3}{*}{\shortstack{\small{Two-branch AE with TSR}}}  & \Checkmark &\Checkmark &\Checkmark & & & & & &48.91&48.22&62.67&63.16\\ 
 &  \Checkmark &\Checkmark &\Checkmark &\Checkmark& & & & &49.02&48.69&62.81&63.47\\ 
 & \Checkmark &\Checkmark &\Checkmark &\Checkmark&\Checkmark & & & &49.01&48.93&62.94&63.42\\
\cmidrule(r){1-13}
\multirow{3}{*}{ \shortstack{\small{Two-branch VAE with TSR} \\ \small{(one standard Gaussian for IAI branch)}}} & \Checkmark &\Checkmark &\Checkmark & & &\Checkmark & & &54.43&53.83&68.20&67.92\\ 
 & \Checkmark &\Checkmark &\Checkmark &\Checkmark & &\Checkmark & & &55.97&54.21&68.74&68.83\\
 & \Checkmark &\Checkmark &\Checkmark &\Checkmark &\Checkmark &\Checkmark & & &56.65&55.82&69.13&69.01\\ \cmidrule(r){1-13}
\shortstack{\small{Two-branch VAE with TSR } \\ \small{(two standard Gaussians for two branches)}}& \Checkmark &\Checkmark &\Checkmark &\Checkmark &\Checkmark &\Checkmark & & &\underline{57.08}&\underline{56.46}&\underline{69.84}&\underline{69.75}\\ \cmidrule(r){1-13}
\multirow{2}{*}{\shortstack{\small{DG-VAE with TSR}\\ \small{(one standard Gaussian and one MoG)}}}  & \Checkmark &\Checkmark &\Checkmark &\Checkmark &\Checkmark &\Checkmark &\Checkmark & &56.13&55.27&68.49&68.20\\ 
 & \Checkmark &\Checkmark &\Checkmark &\Checkmark &\Checkmark &\Checkmark &\Checkmark &\Checkmark&\textbf{59.49}&\textbf{58.46}&\textbf{72.97}&\textbf{71.78}\\    \bottomrule[1pt]
\end{tabular}
\vspace{-1.0em}
\end{table*}

\par
\textbf{Hyperparameter.\label{exp:hyper}} Following the parameters settings in \cite{yu2019robust} and \cite{dai2018cross}, we set $\lambda_{aux}=0.1$ and $m=0.1$. 
In the first stage, we empirically find that training with a large value of $\lambda_{ambi}$ is unstable. We thus set $\lambda_{ambi}$ to $0.001$ in the second stage, and increase it to $0.01$ in the third stage to regularize the disentanglement. We fix $\lambda_{id}$, $\lambda_{cmtl}$, $\lambda_{idc}$ and $\lambda_{cyc}$ to $1$, $1$, $0.5$ and $0.5$, respectively. For other parameters, we fix the split of RegBD dataset and use corresponding images as training/validation sets. We use a grid search on the validation split to set the parameters, resulting in $\lambda_{gmm} = 1$, $\lambda_{lmc} = 0.1$, $\lambda_{disc} = 0.2$ and $\lambda_{rec}= 0.5$. We fix all parameters on both datasets.

\vspace{-1.0em}
\subsection{Comparison with State-of-the-art Methods}

Our proposed DG-VAE outperforms the four types of state-of-the-art VI-ReID methods (see Table~\ref{tab:state}). On the one hand, DG-VAE could be treated as a distribution alignment method since we enforce both modalities to follow the same priors. On the other hand, our proposed method indeed contains the encoder-decoder architecture but dose not generate original images. Hence, we analyze two of the most related types of methods as follows.
\par
\textbf{Image Generation.} Both D2RL\cite{wang2019learning} and JSIA\cite{wang2020cross} employ image stylization networks to generate cross-modality images by predicting the parameters of the AdaIN layers~\cite{DBLP:conf/iccv/HuangB17}, which are shown to mainly captures the style information of the image. However, the visible-infrared discrepancy is more complex than the variation of style characteristics, \textit{e.g.}, unaligned pose variation especially in the more challenging SYSU-MM01 dataset. In contrast, our DG-VAE and AlignGAN\cite{wang2019rgb} utilize the encoder-decoder network without AdaIN layer to model such an intractable situation and reach a better performance. However, without a prior assumption, AlignGAN 
directly maps the learned latent codes to generated images, which leads model to learn only a one-to-one mapping. The AlignGAN struggles to generate cross-modality images when it encounters the unseen inputs in test set, thereby degrading the generalization. Unlike the above methods, DG-VAE simples latent codes from the proposed dual Guassian priors to reconstruct cross-modality feature maps for disentangling IDI and IAI, which results in the significant improvements compared with AlignGAN~\cite{wang2019rgb}, rank-1 accuracy by 16.67\% and mAP by 18.38\% on the RegDB dataset.
\par
\textbf{Distribution Alignment.} The DFE \cite{hao2019dual} achieves the most competitive results, which indicates that bridging the cross-modality gap by decreasing the distribution divergence is effective. However, they estimated a bias distribution drawn from only the observed data. Benefiting from the reparametrization trick, DG-VAE takes additional latent data from re-sampling operation into account, which allows model to explore unobserved data and approximates the estimated distribution with true distribution. Hence, our DG-VAE outperforms DFE~\cite{hao2019dual} in terms of rank-1 accuracy by 10.78\% and mAP by 9.87\% on the SYSU-MM01 dataset. 
\begin{figure*}[th]
  \centering
  \includegraphics[width=\textwidth]{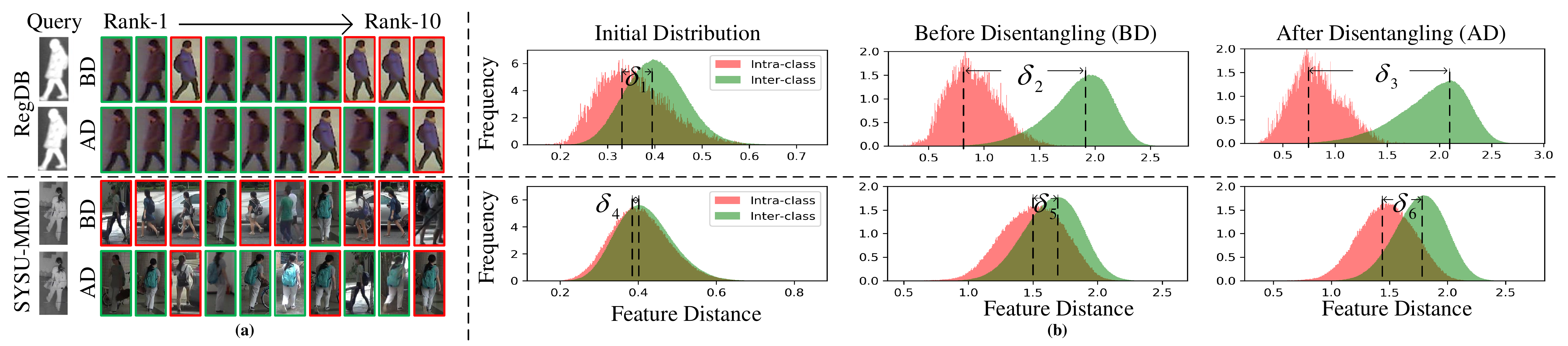}
  \captionsetup{belowskip=-10pt,aboveskip=-2pt}
  \caption{\textbf{(a)} is the illustration of the qualitative examples. The green borders donate the correct matches corresponding the given query and the red is opposite. \textbf{(b)} depicts the distributions of the Euclidean distance between cross-modality (RGB-IR) features. The intra-class and inter-class distances are indicated by \textcolor{red}{red} and \textcolor{green}{green} color, respectively.\label{pic:distance}}
\end{figure*}

\vspace{-1.0em}
\subsection{Ablation Study 
}\label{subsec:ablation}
The first row in Table~\ref{tab:ablation} is the baseline model composed of $\mathit{E}^{V}$, $\mathit{E}^{I}$ and $\mathit{E}^{D}_{\phi}$, which is optimized by $\mathcal{L}_{task}$ with $\lambda_{aux}=0$.
\par
\textbf{Impact of ALMC.} To verify the effectiveness of ALMC, we reset $\lambda_{aux}$ to 0.1. The first three rows indicate that adding only the MoG prior term $\mathcal{L}_{gmm}$ to the baseline model will degrade performance. It is because at the beginning the distribution of each cluster is not stable. Meanwhile, each initial Gaussian component is stochastic, which has a high probability of becoming degenerated. Therefore, when we impose the ALMC $\mathcal{L}_{lmc}$ to collaboratively optimize the model, the performance on both datasets increases with a considerable gain.
\par
\textbf{Impact of TSR.} To demonstrate the effectiveness of TSR, based on the baseline model, we further build a two-branch AE by adding the decoder $\mathit{G}_{\theta}$ and the IAI encoder $\mathit{E}^{A}_{\psi}$ and directly use the separate means of IDI and IDI codes for reconstruction without reparameterization operation. The fourth row shows that the TSR strategy improves the performance of baseline even with the general AE architecture. However, applying the identity and cycle consistent constraints ($\mathit{\mathcal{L}_{idc}}$ and $\mathit{\mathcal{L}_{cyc}}$) is not efficient in such AE. This is because without the augmented samples from the re-sampling operation, the identity-consistent condition is relatively easy to achieve by the reconstruction loss $L_{rec}$.
\par
\textbf{Impact of dual Gaussian priors.} To comprehensively explore the advantages of the proposed dual Gaussian priors, we conduct three series of experiments to verify the performances of two-branch VAEs with one standard Gaussian, two standard Gaussian and the combining of one MoG and one standard Gaussian priors. First, we employ our full model while only executing re-sampling operation on IAI branch. The seventh row shows that such architecture with TSR still obtains improvements on two datasets. The eighth and ninth rows indicate that applying the $\mathit{\mathcal{L}_{idc}}$ and $\mathit{\mathcal{L}_{cyc}}$ further boosts the VI-ReID performance with the VAE architecture. Second, we adopt full model architecture but calculate $\mathcal{L}_{ambi}$ on both branches with re-sampling operation. The eleventh row shows using two standard Gaussian priors obtains mere improvement. Although restricting the IDI following standard Gaussian distribution is beneficial for learning compacted representations, only a standard Gaussian distribution is hard to model the complex multi-cluster structures. Finally, the last two rows demonstrate that our MoG prior effectively alleviates this drawback. It implies that the MoG prior and the triplet swap disentangling are complementary, and combining all the loss terms produces the best results.

\begin{table}[th]
\centering
\vspace{0.5em}
  \captionsetup{aboveskip=4pt}
  
\caption{Comparison of different reconstruction strategies.\label{tab:gan}}
\begin{tabular}{ccccc}
\toprule[1pt]
Datasets & \multicolumn{2}{c}{RegDB} & \multicolumn{2}{c}{Decoder \& Discriminator} \\ \cmidrule(r){1-1} \cmidrule(r){2-3} \cmidrule(r){4-5}
Strategies     & R-1         & mAP         & Params(M)     & FLOPs(G)    \\ 
 \midrule[1pt]
Feature Maps   &72.97&71.78&16.26&1.23\\
\textit{with discriminator} & 73.41 & 72.09 &18.52&1.50\\ \hline
Original Images & 72.62       & 72.15      &31.22&6.03\\
\textit{with discriminator}& \textbf{73.74}&\textbf{72.85}&47.10&6.56\\ 
\bottomrule[1pt]
\end{tabular}
\vspace{-0.5em}
\end{table}


\vspace{-1.0em}
\subsection{Discussion}\label{subsec:discussion}
\par
\textbf{Why are MoG prior term and ALMC efficient?} One advantage of the DG-VAE is to represent feature embedding as a distribution instead of a fixed feature vector. When our model encounters the outliers caused by noise label and occlusion as shown in Appendix \ref{append:dataset}, it will assign the larger variance to these samples instead of sacrificing inter-class separability, thereby promoting robust feature learning. In addition, the last three rows of Table~\ref{tab:ablation} suggest that the MoG prior is more beneficial for feature disentanglement than feature learning.
As for feature disentanglement, class-specific prior leads the model to explicitly handle the variation of each identity, which is equivalent to extracting IDI across different modalities. It allows this kind of information to flow into the IDI branch, thereby promoting the disentangling process. On the other hand, a more reasonable distribution of latent variates enables more realistic yields from the generator. In short, with the ALMC, the MoG prior makes extracting the disentangled features more efficient.

\par
\textbf{Why is triplet swap disentangling efficient for VI-ReID?} The retrieval features learned by most existing VI-ReID methods are often highly entangled representations. Intuitively, cross-modality tasks suffer from more data discrepancies than single-modality tasks. Thus, excluding IAI plays a crucial role in boosting performance. The qualitative examples are shown in Fig.~\ref{pic:distance}\textbf{(a)}. Another reason for its efficiency becomes clear if we count the frequency of inter-class and intra-class distance as illustrated in Fig.~\ref{pic:distance}\textbf{(b)}. Comparing the first and second columns, the means of inter- and intra-distance are pushed away by using the baseline method with the MoG prior, where $\delta_{1}<\delta_{2}$ and $\delta{4}<\delta{5}$. Note that due to the difficulty of the SYSU-MM01 dataset, where $\delta{4}\approx 0$, it is hard to significantly push two peaks away like for the RegDB dataset, where $\delta{2}>\delta{5}$. Thanks to disentangling ID-discriminable factors, our full model performs considerably better on both datasets, $\delta{3}\approx 1.4$ and $\delta{6}\approx 0.4$, which means 13.05\% and 10.86\% improvement of mAP on the SYSU-MM01 and RegDB datasets, respectively. As a consequence, based on factorized IDI, DG-VAE indeed leads the intra-class embeddings to be compacted and disperses the inter-class clusters with a large margin as illustrated in Appendix~\ref{app:tsne}, thereby achieving better performance.

\par
\textbf{Why does DG-VAE reconstruct the feature map instead of the original image?} We experiment with different settings as shown in Table~\ref{tab:gan}. Our full DG-VAE model is regarded as baseline setting in the first row. As shown in the second row, for a fair comparison, we treat the reconstructed feature maps as a multi-channel image and add an advanced PatchGAN discriminator \cite{isola2017image} to distinguish whether the generated multi-channel image is fake. Moreover, we add four extra deconvolutional layers to generate fake images with the same height and width as the original image, but don't calculate binary cross-entropy loss function of the discriminator. The difference between the third and fourth rows is whether the binary cross-entropy loss function of the discriminator is calculated or not. The results show that our feature maps reconstruction strategy gets a competitive performance compared to generating the original image. Meanwhile, our model requires only half of the parameters and one-fifth FLOPs. We find increased performance can be attributed to the extra discriminator with binary cross-entropy loss function. Nevertheless, we do not have the motivation to distinguish reconstructed feature maps. On the other hand, DG-VAE starts from the dynamically learned feature maps instead of the fixed input images, which is different from the assumption of Bayes evidential reasoning.

\vspace{-0.5em}
\section{CONCLUSION}
\par
In this paper, we design a novel variational disentanglement architecture with dual Gaussian constraints and triplet swap reconstruction for robust cross-modality VI-ReID. Our model learns two separate encoders with dual Gaussian priors and factorizes the latent variate into ID-discriminable and ID-ambiguous codes. By excluding ID-ambiguous information, we show that the learned representations achieve significant improvement compared to the baseline and reaches a competitive performance with state-of-the-art methods as well. Furthermore, we study the different reconstruction strategies, analyze the reason for the observed increase in performance, and discuss the principle of our derived MoG prior regularizer. The proposed method shows considerable potential of the disentangled representation for multi-modality tasks.

\clearpage
\bibliographystyle{ACM-Reference-Format}
\normalem
\bibliography{egbib}
\clearpage
\onecolumn
\section*{APPENDIX}
\appendix
\section{The Derivation of our ELBO}\label{app:ELBO}
\par
Here we provide the derivation of our ELBO. The basic probabilities are defined as:
\begin{equation}
p(y)=Cat(\boldsymbol{\pi}),
\end{equation}
\begin{equation}
q_{\psi}(\mathbf{a}|\mathbf{f})=\mathcal{N}(\mathbf{a};\boldsymbol{\mu}_{\psi(\mathbf{f})}, \boldsymbol{\sigma}_{\psi(\mathbf{f})}^{2}),
\end{equation}
where $Cat(\boldsymbol{\pi})$ is the categorical distribution parametrized by $\boldsymbol{\pi}$, $\boldsymbol{\pi}\in \mathbb{R}_{+}^N$, $1=\sum_{y=1}^{N}\boldsymbol{\pi}_{y}$. $p(y)=\boldsymbol{\pi}_{y}$ is the prior probability for identity $y$, which is simply set to $\frac{1}{N}$ for all categories. $q_{\psi}(\mathbf{a}|\mathbf{f})$ is the encoder $\mathit{E}_{\psi}^{A}$ whose input is $\mathbf{f}$ and is parametrized by $\psi$. $\mathcal{N}(\mathbf{a};\boldsymbol{\mu}_{\psi(\mathbf{f})}, \boldsymbol{\sigma}_{\psi(\mathbf{f})}^{2})$ are multivariate Gaussian distribution parametrized by $\boldsymbol{\mu}_{\psi(\mathbf{f}}$ and $\boldsymbol{\sigma}_{\psi(\mathbf{f})}$.
\par
Since the approximate posterior $q(\mathbf{d},y|\mathbf{f})$ is intractable, we assume conditional dependence between $\mathbf{d}$ and the identity $y$. Thus, it could be factorized as following form:
\begin{equation}
q(\mathbf{d},y|\mathbf{f}) = q(y|\mathbf{d})q_{\phi}(\mathbf{d}|\mathbf{f}).
\end{equation}
\par
To handle the multiple clusters of data in a latent space, we expect that the ID-discriminable codes $\mathbf{d}$ follow the MoG prior distribution $p(\mathbf{d})$ with class-specific mean and unit variance, where each component corresponds to a particular identity $y$, and $p(y)$ is the prior probability, which is simply set to $\dfrac{1}{N_{y}}$ for all identities. The variance of each component models the intra-identity variations. Thus, we assume that the prior probability $p(\mathbf{d})$ is as follows,
\begin{equation}
p(\mathbf{d}) = \sum_{y}p(y)\mathcal{N}(\mathbf{d};\boldsymbol{\mu}_{y}, \mathbf{I}).
\end{equation}
\par
Naturally, the conditional $q(\mathbf{d}|y)$ could be derived to the following form,
\begin{equation}
p(\mathbf{d}|y) = \mathcal{N}(\mathbf{d};\boldsymbol{\mu}_{y}, \mathbf{I}).
\end{equation}
\par
Then, we employ a encoder $\mathit{E}_{\phi}^{D}$ whose input is $\mathbf{f}$ and is parametrized by $\phi$ to fit the prior distribution $p(\mathbf{d})$, which is formulated as follows,
\begin{equation}
q_{\phi}(\mathbf{d}|\mathbf{f})= \mathcal{N}(\mathbf{d};\boldsymbol{\mu}_{\phi(\mathbf{f})}, \boldsymbol{\sigma}_{\phi(\mathbf{f})}^{2}).
\end{equation}
\par
In this paper, we not only take the generative model $p_{\theta}(\mathbf{\hat{f}}|\mathbf{a},\mathbf{d})$ into account, but also consider two inference models $q_{\psi}(\mathbf{a}|\mathbf{f})$ and $q(\mathbf{d},y|\mathbf{f})$ for two branches, respectively. The probabilistic graphical models are illustrated in Fig.~\ref{pic:pgm}.
\par
Firstly, we suppose the \textbf{inference process} as follows: Given the feature maps $\mathbf{f}$ extracted by two modality-specific feature extractors, we simultaneously feed them into the IDI encoder $p_{\phi}(\mathbf{d}|\mathbf{f})$ and IAI encoder $p_{\psi}(\mathbf{a}|\mathbf{f})$, which results in the ID-discriminable code $\mathbf{d}$ and ID-ambiguous code $\mathbf{a}$ conditionally depending on the extracted feature maps, \textit{i.e.}, $\mathbf{d} \sim p_{\phi}(\mathbf{d}|\mathbf{f})$ and $\mathbf{a} \sim p_{\psi}(\mathbf{a}|\mathbf{f})$. Then, we apply a latent code classifier to fit the condition probability $p(y|\mathbf{d})$, which enables our DG-VAE to infer the correct identity for VI-ReID task.
\par
Based on the above inference process, we thus factorize the joint distribution $q(\mathbf{f},y)$ as follows:
\begin{equation}
q(\mathbf{f},\mathbf{a},\mathbf{d},y) = q(y|\mathbf{d})q_{\phi}(\mathbf{d}|\mathbf{f})q_{\psi}(\mathbf{a}|\mathbf{f}).
\end{equation}
\par
Secondly, we suppose the \textbf{generative process} as follows: Give $N_{y}$ identities, a reconstructed feature map $\hat{\mathbf{f}}$ is generated by sampling an ID-ambiguous code $\mathbf{a}$ from $p(\mathbf{a})$ and a corresponding identity $y$, then, an ID-discriminable code $\mathbf{d}$ is sampled from the conditional distribution $\mathbf{d}\sim p(\mathbf{d}|y)$. Finally, a decoder $p_{\theta}(\hat{\mathbf{f}}|\mathbf{d},\mathbf{a})$ maps the combination of $\mathbf{a}$ and $\mathbf{d}$ to a reconstructed feature map $\hat{\mathbf{f}}$.
\par 
According to the above generative process, we factorize the joint distribution $p_{\theta}(\hat{\mathbf{f}},\mathbf{d},\mathbf{a},y)$ as:
\begin{equation}\label{eq:likelihood}
p(\hat{\mathbf{f}},\mathbf{a},\mathbf{d},y)=p_{\theta}(\hat{\mathbf{f}}|\mathbf{a},\mathbf{d})p(\mathbf{a})p(\mathbf{d}|y)p(y).\\
\end{equation}
\par
Finally, by using Jensen's inequality, the log-likelihood $\log p(\mathbf{\hat{f}})$ can be written as:
\begin{equation}\label{eq:elboproof}
\begin{split}
\log &p(\hat{\mathbf{f}}) =\log\iint\sum_{y}p_{\theta}(\hat{\mathbf{f}}|\mathbf{a},\mathbf{d})p(\mathbf{a})p(\mathbf{d}|y)p(y)d\mathbf{a}d\mathbf{d}\\
& =\log \iint\sum_{y}q(y|\mathbf{d})q_{\phi}(\mathbf{d}|\mathbf{f})q_{\psi }(\mathbf{a}|\mathbf{f}) \frac{p_{\theta}(\hat{\mathbf{f}}|\mathbf{a},\mathbf{d})p(\mathbf{a})p(\mathbf{d}|y)p(y)}{q(y|\mathbf{d})q_{\phi}(\mathbf{d}|\mathbf{f})q_{\psi}(\mathbf{a}|\mathbf{f})}d\mathbf{a}d\mathbf{d} \\
&=\log \mathbb{E}_{y\sim q(y|\mathbf{d}),\mathbf{d}\sim q_{\phi}(\mathbf{d}|\mathbf{f}),\mathbf{a}\sim q_{\psi }(\mathbf{a}|\mathbf{f})} \frac{p_{\theta}(\hat{\mathbf{f}}|\mathbf{a},\mathbf{d})p(\mathbf{a})p(\mathbf{d}|y)p(y)}{q(y|\mathbf{d})q_{\phi}(\mathbf{d}|\mathbf{f})q_{\psi }(\mathbf{a}|\mathbf{f})}\\
& \geq \mathbb{E}_{y\sim q(y|\mathbf{d}),\mathbf{d}\sim q_{\phi}(\mathbf{d}|\mathbf{f}),\mathbf{a}\sim q_{\psi }(\mathbf{a}|\mathbf{f})} [\log \frac{p_{\theta}(\hat{\mathbf{f}}|\mathbf{a},\mathbf{d})p(\mathbf{a})p(\mathbf{d}|y)p(y)}{q(y|\mathbf{d})q_{\phi}(\mathbf{d}|\mathbf{f})q_{\psi }(\mathbf{a}|\mathbf{f})}]\\
& = \mathbb{E}_{\mathbf{d}\sim q_{\phi}(\mathbf{d}|\mathbf{f}),\mathbf{a}\sim q_{\psi }(\mathbf{a}|\mathbf{f})}[\log p_{\theta} (\hat{\mathbf{f}}|\mathbf{d},\mathbf{a})]\\
& \quad + \mathbb{E}_{\mathbf{d}\sim q_{\phi}(\mathbf{d}|\mathbf{f}),\mathbf{a}\sim q_{\psi }(\mathbf{a}|\mathbf{f})}[\log \frac{p(\mathbf{a})}{q_{\psi}(\mathbf{a}|\mathbf{f})}]\\
& \quad + \sum_{y}q(y|\mathbf{d})\mathbb{E}_{\mathbf{d}\sim q_{\phi}(\mathbf{d}|\mathbf{f}),\mathbf{a}\sim q_{\psi }(\mathbf{a}|\mathbf{f})}[\log\frac{p(\mathbf{d}|y)}{q_{\phi}(\mathbf{d}|\mathbf{f})}]\\
& \quad + \mathbb{E}_{\mathbf{d}\sim q_{\phi}(\mathbf{d}|\mathbf{f}),\mathbf{a}\sim q_{\psi }(\mathbf{a}|\mathbf{f})}[\sum_{y}q(y|\mathbf{d})\log\frac{p(y)}{q(y|\mathbf{d})}]\\
& = \mathbb{E}_{\mathbf{d}\sim q_{\phi}(\mathbf{d}|\mathbf{f}),\mathbf{a}\sim q_{\psi }(\mathbf{a}|\mathbf{f})} [\log p_{\theta} (\hat{\mathbf{f}}|\mathbf{d},\mathbf{a})]\\
& \quad - D_{KL}(q_{\psi}(\mathbf{a}|\mathbf{f})||p(\mathbf{a}))\\
& \quad - \sum_{y}q(y|\mathbf{d})D_{KL}(q_{\phi}(\mathbf{d}|\mathbf{f})||p(\mathbf{d}|y))\\
& \quad - D_{KL}(q(y|\mathbf{d})||p(y))\\
& = - D_{KL}(q_{\phi}(\mathbf{d},y|\mathbf{f})||p(\mathbf{d},y))\\
& \quad - D_{KL}(q_{\psi}(\mathbf{a}|\mathbf{f})||p(\mathbf{a}))\\
& \quad + \mathbb{E}_{q_{\phi}(\mathbf{d},y|\mathbf{f}),q_{\psi}(\mathbf{a}|\mathbf{f})} [\log p_{\theta} (\hat{\mathbf{f}}|\mathbf{d},\mathbf{a})],\\
\end{split}
\end{equation}
\par
where $D_{KL}$ denotes Kullback-Leibler divergence, \textit{i.e.}, $D_{KL}=-\int p(z)\frac{p(z)}{q(z)}dz$. $\pi_{y}$ is the prior probability for identity $y$, $\pi\in \mathbb{R}_{+}^N$, $1=\sum_{y=1}^{N}\pi_{y}$, $Cat(\boldsymbol{\pi})$. The result of Eq. \ref{eq:elboproof} corresponds to the above mentioned our ELBO in Eq.\ref{eq:elbo}.

\clearpage
\section{The Derivation of MoG Prior Term}\label{app:gmm}
Due to the non-negative property of KL divergence, maximizing ELBO is equivalent to minimizing the KL divergence between approximate posterior and prior distribution --- a so-called ``Gaussian mixture prior regularization term''. By using reparametrization tricks \cite{KingmaW13}, the regularization term could be calculated as follows,
\begin{equation}\label{eq:gmprt}
\begin{split}
&\quad \sum_{y}q(y|\mathbf{d})D_{KL}(q_{\phi}(\mathbf{d}|\mathbf{f})||p(\mathbf{d}|y))\\
&=\sum_{y}q(y|\mathbf{d})\mathbb{E}_{\mathbf{d}\sim q_{\phi}(\mathbf{d}|\mathbf{f}),\mathbf{a}\sim q_{\psi }(\mathbf{a}|\mathbf{f})}[\log\frac{p(\mathbf{d}|y)}{q_{\phi}(\mathbf{d}|\mathbf{f})}]\\
&=\mathbb{E}_{\mathbf{d}\sim q_{\phi}(\mathbf{d}|\mathbf{f}),\mathbf{a}\sim q_{\psi }(\mathbf{a}|\mathbf{f})}\sum_{y}q(y|\mathbf{d})[\log\frac{p(\mathbf{d}|y)}{q_{\phi}(\mathbf{d}|\mathbf{f})}]\\
&=\mathbb{E}_{\mathbf{d}\sim q_{\phi}(\mathbf{d}|\mathbf{f}),\mathbf{a}\sim q_{\psi }(\mathbf{a}|\mathbf{f})}\sum_{c}\mathbbm{1}(y=c)\log\frac{p(\mathbf{d}|y)}{q_{\phi}(\mathbf{d}|\mathbf{f})},\\
\end{split}
\end{equation}
\par 
where the conditional probability $q(y|\mathbf{d})$ in supervised setting is the one-hot encoding of the label $y$. Then, according to Eq.~\ref{eq:pdy} and Eq.~\ref{eq:pdf}, we simplify the inner expectation term as follows,
\begin{equation}
\begin{split}
& \quad \log(\frac{p(\mathbf{d}|y)}{q_{\phi}(\mathbf{d}|\mathbf{f})})\\
& = \ln \Bigg( \frac{\frac{1}{(2\pi)^{l/2}} \exp\left\{-\frac{1}{2} \left \| \mathbf{d} - \boldsymbol{\mu}_{y}\right \|^2\right\}}{\frac{1}{(2\pi)^{l/2}det(\boldsymbol{\sigma}_{\phi(\mathbf{f})})}\exp\left\{-\frac{1}{2}\left\Vert\frac{\mathbf{d} - \boldsymbol{\mu}_{\phi(\mathbf{f})}}{\boldsymbol{\sigma}_{\phi(\mathbf{f})}}\right\Vert^2\right\}} \Bigg)\\
& = \ln \big(det(\boldsymbol{\sigma}_{\phi(\mathbf{f})})\big) - \frac{1}{2}  \left \| \frac{\boldsymbol{\sigma}_{\phi(\mathbf{f})}} {\mathbf{d} - \boldsymbol{\mu}_{\phi(\mathbf{f})}} \right \| ^{2} + \frac{1}{2} \left \| \mathbf{d} - \boldsymbol{\mu}_{y}\right \| ^{2},\\
\end{split}
\end{equation}
where $l$ is the dimensionality of $\boldsymbol{\sigma}_{\phi(\mathbf{f})}$. According to the reparametrization trick, the second term could be regard as an unrelated variate. Thus, the Gaussian mixture prior regularization term in Eq.~\ref{eq:gmprt} can be derive to a closed solution as follows,
\begin{equation}
\begin{split}
&\quad \mathbb{E}_{\mathbf{d}\sim q_{\phi}(\mathbf{d}|\mathbf{f}),\mathbf{a}\sim q_{\psi }(\mathbf{a}|\mathbf{f})}\sum_{c}\mathbbm{1}(y=c)\log\frac{p(\mathbf{d}|y)}{q_{\phi}(\mathbf{d}|\mathbf{f})}\\
&= \mathbb{E}_{\mathbf{d}\sim q_{\phi}(\mathbf{d}|\mathbf{f}),\mathbf{a}\sim q_{\psi }(\mathbf{a}|\mathbf{f})}\sum_{c}\mathbbm{1}(y=c)\Big(\ln \big(det(\boldsymbol{\sigma}_{\phi(\mathbf{f})})\big) + \frac{1}{2} \left \| \mathbf{d} - \boldsymbol{\mu}_{y}\right \| ^{2}\Big).\\
\end{split}
\end{equation}

\clearpage
\section{The Proof and Derivation of maximum entropy regularizer}\label{app:mr}
We factorize the final term in Eq.~\ref{eq:elboproof} as follows,
\begin{equation}
\begin{split}
\quad & D_{KL}(q(y|\mathbf{d})||p(y))\\
& = \sum_y q(y|\mathbf{d})\log \frac{q(y|\mathbf{d})}{p(y)}\\
& = \sum_y q(y|\mathbf{d})\log q(y|\mathbf{d}) - \sum_{y} q(y|\mathbf{d})\log p(y)\\
& = -\mathcal{H}(y|\mathbf{d}) - constant.\\
\end{split}
\end{equation}
\par
Note that we assume the prior distribution $p(y)$ is uniform distribution so that the second term is a constant. And the first term is explicit negative entropy of conditional probability $q(y|\mathbf{d})$. Minimizing the KL divergence between $q(y|\mathbf{d})$ and $p(y)$ results in maximizing conditional entropy, $\mathcal{H}(y|\mathbf{d})$. Combining the Gaussian mixture prior regularization term, this is indeed performing an unsupervised clustering algorithm. It expects that the posterior distribution align the corresponding Gaussian component while the maximizing entropy encourage all Gaussian components to be non-overlapping, thus preventing the so-called ``posterior collapse''. To accomplish the same goal in supervised setting, we impose an adaptive large-margin constraint for each Gaussian component inspired by \cite{liu2019adaptiveface}. Given some samples following Mixture-of-Gaussians (MoG) distribution, the samples lie on a MoG manifold. The distance between a sample $\mathbf{d}$ with identity $y$ and mean of corresponding component $\boldsymbol{\mu}_{y}$ should be define as squared Mahalanobis distance:
\begin{equation}\label{app:result:gmprt}
\mathcal{L}_{lmc} = - \mathbb{E}_{\mathbf{d}\sim q_{\phi}(\mathbf{d}|\mathbf{f})}\log \dfrac{\exp^{-\mathit{D}_{M}(\mathbf{d},y,\mathbf{f})}}{\sum_{c=1,c\neq y}^{N_{y}}\exp^{-\left \|\mathbf{d}-\boldsymbol{\mu}_{c}\right\|^{2}} + \exp^{-\mathit{D}_{M}(\mathbf{d},y,\mathbf{f})}},
\end{equation}
where the indicator function $\mathbbm{1}()$ equals $1$ if $y$ equals $k$; and $0$ otherwise. Furthermore, given some samples following Mixture-of-Gaussians (MoG) distribution, the samples lie on the MoG manifold. The distance between a sample $\mathbf{d}$ with identity $y$ and the corresponding mean $\boldsymbol{\mu}_{\phi(\mathbf{f})}$ should be defined as the squared Mahalanobis distance $\mathit{D}_{M}(\mathbf{d},y,\mathbf{f})$:
\begin{equation}\label{app:margin}
\mathit{D}_{M}(\mathbf{d},y,\mathbf{f}) = (\mathbf{d} - \boldsymbol{\mu}_{\phi(\mathbf{f})})^{T}\boldsymbol{\sigma}_{\phi(\mathbf{f})}^{-1}(\mathbf{d} - \boldsymbol{\mu}_{\phi(\mathbf{f})}) - \alpha_{y},\\
\end{equation} 
where the identity-specific margin $\alpha\geq 0$ is a learnable parameter. This is so-called adaptive large-margin MoG constraint. 

\clearpage
\section{The conception of an information theory perspective for Our DG-VAE}\label{app:itp}
\par
To tackle these two co-existing problems, we rethink the VI-ReID task from a perspective of mutual information \cite{DBLP:conf/iclr/AlemiFD017}. The core in VI-ReID is to maximize the mutual information among the latent representations of visible-infrared images and labels, $\mathcal{I}(Z^{V};Z^{I};Y)$, which allows the learned representation to server as the clues to establish cross-modality connections. We call this kind of information as the cross-modality \textit{ID-discriminable information} (IDI), \textit{e.g.}, the shape and outline of a body, and some latent characteristics. However, a general method delivers this goal by respectively maximizing the mutual information between labels and the different-modality latent representations, \textit{i.e.},  $\mathcal{I}(Z^{V};Y)$ and $\mathcal{I}(Z^{I};Y)$. It leads to exist substantial information, $\mathcal{I}(Z^{I};Y|Z^{V})$ and $\mathcal{I}(Z^{V};Y|Z^{I})$, only belonging to each modality, \textit{e.g.}, color information for visible images, and thermal information for infrared images. We call this kind of information as the cross-modality \textit{ID-ambiguous information} (IAI). If we perform retrieval by directly calculating the $L_{2}$ distance between two such vectors, the performance of retrieval feature will be amortized due to noise caused by the redundant IAI dimensions, as illustrated in Fig.~\ref{pic:main1b}.
\begin{figure*}[h]
  \centering
  \includegraphics[width=\textwidth]{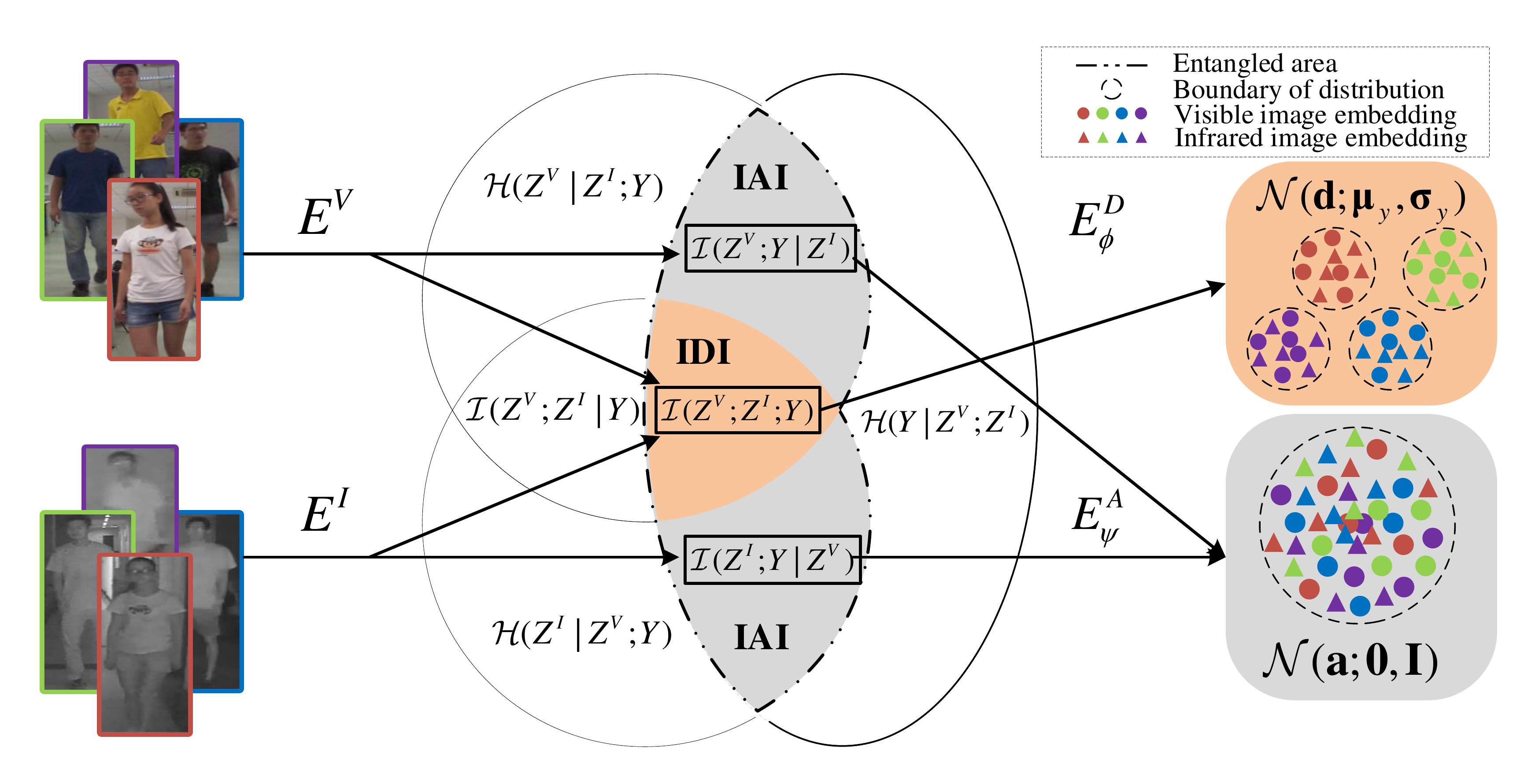}
  \caption{Illustration of our information theory perspective, where $\mathcal{I}(Z^{V};Z^{I};Y)$ is multi-variate mutual information. $\mathcal{H}Z^{V}|Z^{I};Y$ and $\mathcal{H}Z^{I}|Z^{V};Y$ denote multi-variate conditional entropy, which represent the uncertainty of $\mathcal{H}Z^{V}$ and $\mathcal{H}Z^{I}$ under the condition of known $Z^{I};Y$ and $Z^{V};Y$, respectively. Maximizing $\mathcal{I}(Z^{V};Z^{I};Y)$ encourages our model to extract the only modality-shared information corresponding to ground-truth label instead of modality-separated information corresponding to label. Nevertheless, a class of general methods achieve this goal by respectively maximizing the mutual information between labels and the different-modality inputs, \textit{i.e.},  $\mathcal{I}(Z^{V};Y)$ and $\mathcal{I}(Z^{I};Y)$. It indeed leads to extract substantial useless information, $\mathcal{I}(Z^{I};Y|Z^{V})$ and $\mathcal{I}(Z^{V};Y|Z^{I})$, only belonging to each modality.}
\end{figure*}

\clearpage
\section{Connect to Distribution Net}\label{append:dataset}
\begin{figure*}[!h]
  \centering
  \includegraphics[width=\textwidth]{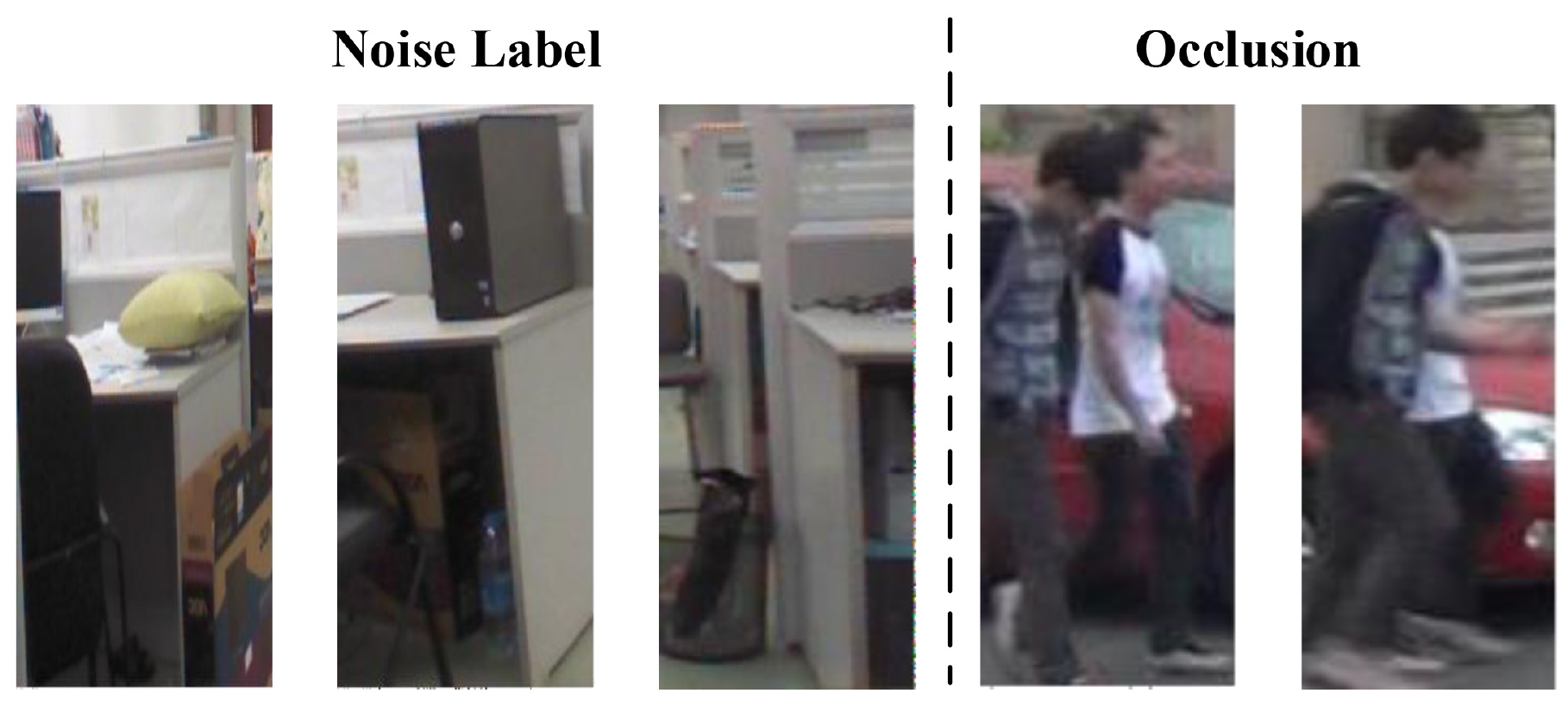}
  \caption{The illustration of noise label examples and occlusion examples in SYSU-MM01 dataset.\label{pic:dataset}}
\end{figure*}

our proposed DG-VAE could connect with DistributionNet\cite{yu2019robust}, which aims at reducing the influences of noise label or outlier by estimating the uncertainty of features. In this work, Yu \textit{et~al.} suggest that modelling a sample by distribution instead of a fixed feature vector enables the model to be more robust against for noise label and achieve a better capability of generalization. At the same time, we find the SYSU-MM01 dataset includes quite a few noise images, such as the occluded and the wrong labelled, as shown in Fig~\ref{pic:dataset}. Thus, our proposed DG-VAE disentangles IDI and IAI while overcoming the influences of noise data simultaneously, which significantly outperforms other compared methods on the SYSU-MM01 dataset.
\clearpage
\section{The detailed neural network architecture}\label{append:nna}
\par
We illustrate our proposed architectures as follows:
\begin{table*}[h]
\caption{The detailed architectures of RGB and IR Feature Maps Extractors.}
\begin{tabular}{|l|c|c|c|}
\hline
Layer Name & Input Size            & Output Size           & RGB and IR Feature Maps extractors \\ \hline
conv1      &$3\times384\times128$&$64\times192\times64$&$7\times7,64,stride~2$\\ \hline
max pool   & $64\times192\times64$ & $64\times96\times32$ &$3\times3,max-pooling, stride~2$                        \\ \specialrule{0.02em}{3pt}{3pt}
conv2\_1 to conv2\_3   & $64\times96\times32$  & $256\times96\times32$  &  
$
 \left[
 \begin{matrix}
   1\times1,~64 \\
   3\times3,~64 \\
   1\times1,~256
  \end{matrix}
  \right] \times 3
$                   \\ \specialrule{0.02em}{3pt}{3pt}
conv3\_1 to conv3\_4   & $256\times48\times16$  & $512\times24\times8$ &  
$
 \left[
 \begin{matrix}
   1\times1,~128 \\
   3\times3,~128 \\
   1\times1,~512
  \end{matrix}
  \right] \times 4
$
\\ \specialrule{0.02em}{3pt}{3pt}
conv4\_1   & $512\times24\times8$  & $1024\times24\times8$ &  
$
 \left[
 \begin{matrix}
   1\times1,~256 \\
   3\times3,~256 \\
   1\times1,~1024
  \end{matrix}
  \right] \times 1
$
\\ \specialrule{0.02em}{3pt}{3pt}
\end{tabular}
\end{table*}

\begin{table*}[h]
\caption{The detailed architectures of ID-discriminable Encoders.}
\begin{tabular}{|l|c|c|c|}
\hline
Layer Name & Input Size            & Output Size           & IDI Encoder\\ \specialrule{0.02em}{3pt}{3pt}
conv4\_2 to conv4\_6   & $512\times24\times8$  & $1024\times24\times8$ &  
$
 \left[
 \begin{matrix}
   1\times1,~256 \\
   3\times3,~256 \\
   1\times1,~1024
  \end{matrix}
  \right] \times 5
$ 
\\ \specialrule{0.02em}{3pt}{3pt}
conv5\_1 to conv5\_3   & $1024\times24\times8$  & $2048\times12\times4$ &  
$
 \left[
 \begin{matrix}
   1\times1,~512 \\
   3\times3,~512 \\
   1\times1,~2048
  \end{matrix}
  \right] \times 3
$ 
\\ \specialrule{0.02em}{3pt}{3pt}
pooling   & $2048\times12\times4$  & $2048\times1\times1$ &  
$ 12\times4, max-pooling$ \\ \specialrule{0.02em}{3pt}{3pt}
mean prediction layer & $2048\times1\times1$  & $2048$ &  
$
 \left[
 \begin{matrix}
   1\times1,~2048 \\
  \end{matrix}
  \right] \times 1
$ \\ \specialrule{0.02em}{3pt}{3pt}
variance prediction layer  & $2048\times1\times1$  & $2048$ &  
$
 \left[
 \begin{matrix}
   1\times1,~2048 \\
  \end{matrix}
  \right] \times 1
$ \\ \specialrule{0.02em}{3pt}{3pt}
\end{tabular}
\end{table*}

\begin{table*}[h]
\caption{The detailed architectures of ID-ambiguous Encoders.}
\begin{tabular}{|l|c|c|c|}
\hline
Layer Name & Input Size            & Output Size           &  IAI Encoder \\ \specialrule{0.02em}{3pt}{3pt}
conv4\_2 to conv4\_6  & $512\times24\times8$  & $1024\times24\times8$ & $
 \left[
 \begin{matrix}
   1\times1,~256 \\
   3\times3,~256 \\
   1\times1,~1024
  \end{matrix}
  \right] \times 5
$
\\ \specialrule{0.02em}{3pt}{3pt}
conv5\_1 to conv5\_3   & $1024\times24\times8$  & $2048\times12\times4$ & $
 \left[
 \begin{matrix}
   1\times1,~512 \\
   3\times3,~512 \\
   1\times1,~2048
  \end{matrix}
  \right] \times 3
$
\\ \specialrule{0.02em}{3pt}{3pt}
pooling   & $2048\times12\times4$  & $2048\times1\times1$ & $12\times4, average-pooling$ \\ \specialrule{0.02em}{3pt}{3pt}
mean prediction layer   & $2048\times1\times1$  & $512$  & $2048\times512, fully-connection$ \\ \specialrule{0.02em}{3pt}{3pt}
variance prediction layer   & $2048\times1\times1$  & $512$  & $2048\times512, fully-connection$ \\ \specialrule{0.02em}{3pt}{3pt}
\end{tabular}
\end{table*}

\begin{table*}[h]
\caption{The detailed architectures of Decoder.}
\begin{tabular}{|l|c|c|c|}
\hline
Layer Name & Input Size            & Output Size           &  IAI Encoder \\ \specialrule{0.02em}{3pt}{3pt}
fc  & $2560 (2048+512) \times1\times1$ & $512\times1\times1$ & $2560\times512, fully-connection$
\\ \specialrule{0.02em}{3pt}{3pt}
deconv1  & $512\times1\times1$  & $512\times6\times2$ & $
 \left[
 \begin{matrix}
   6\times2,~512\\
   BatchNorm\\
   LeakyReLU\\
   Dropout
  \end{matrix}
  \right] \times 1
$
\\ \specialrule{0.02em}{3pt}{3pt}
deconv2  & $512\times6\times2$  & $512\times12\times4$ & $
 \left[
 \begin{matrix}
   4\times4,~512\\
   BatchNorm\\
   LeakyReLU\\
   Dropout
  \end{matrix}
  \right] \times 1
$
\\ \specialrule{0.02em}{3pt}{3pt}
deconv3  & $512\times12\times4$  & $1024\times24\times8$ & $
 \left[
 \begin{matrix}
   4\times4,~1024\\
   BatchNorm\\
   LeakyReLU\\
  \end{matrix}
  \right] \times 1
$
\\ \specialrule{0.02em}{3pt}{3pt}
activation   & $1024\times24\times8$  & $1024\times24\times8$ & $Tanh$ \\ \specialrule{0.02em}{3pt}{3pt}
\end{tabular}
\end{table*}

\clearpage
\section{The visualization of T-SNE for the RegDB and The SYSU-MM01 datasets}
We utilize T-SNE\cite{maaten2008visualizing} visualization to show the overall distribution of embeddings from the test sets of both datasets. Furthermore, we illustrate how the adaptive large margin Gaussian constraint works on two following cases\label{app:tsne}.
\begin{figure*}[ht]
  \centering
  \includegraphics[width=13cm]{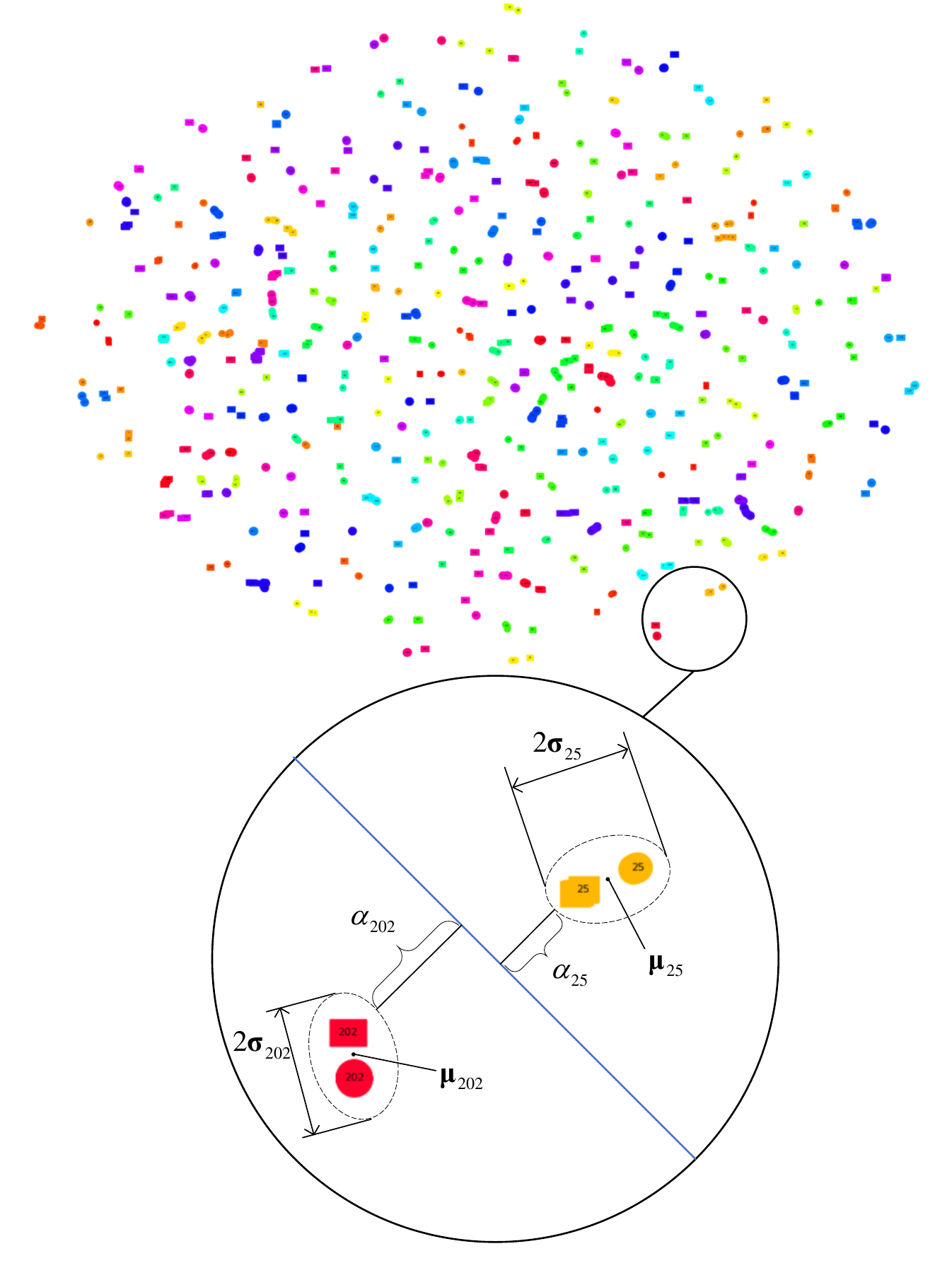}
  \caption{Visualization of embeddings from the test set of RegDB dataset. The circles and squares donate the embeddings from IR and RGB images, respectively. The numbers in the circles and squares are the IDs of embeddings. Different colors represent different IDs. The blue line illustrates hyper-plane boundaries. It is obvious that the heterogeneous embeddings corresponding to same label are superposed. Meanwhile, the inter-class margin is much larger than the intra-class distance cluster by using our DG-VAE method.}
\end{figure*}

\clearpage
\begin{figure*}[ht]
  \centering
  \includegraphics[width=15cm]{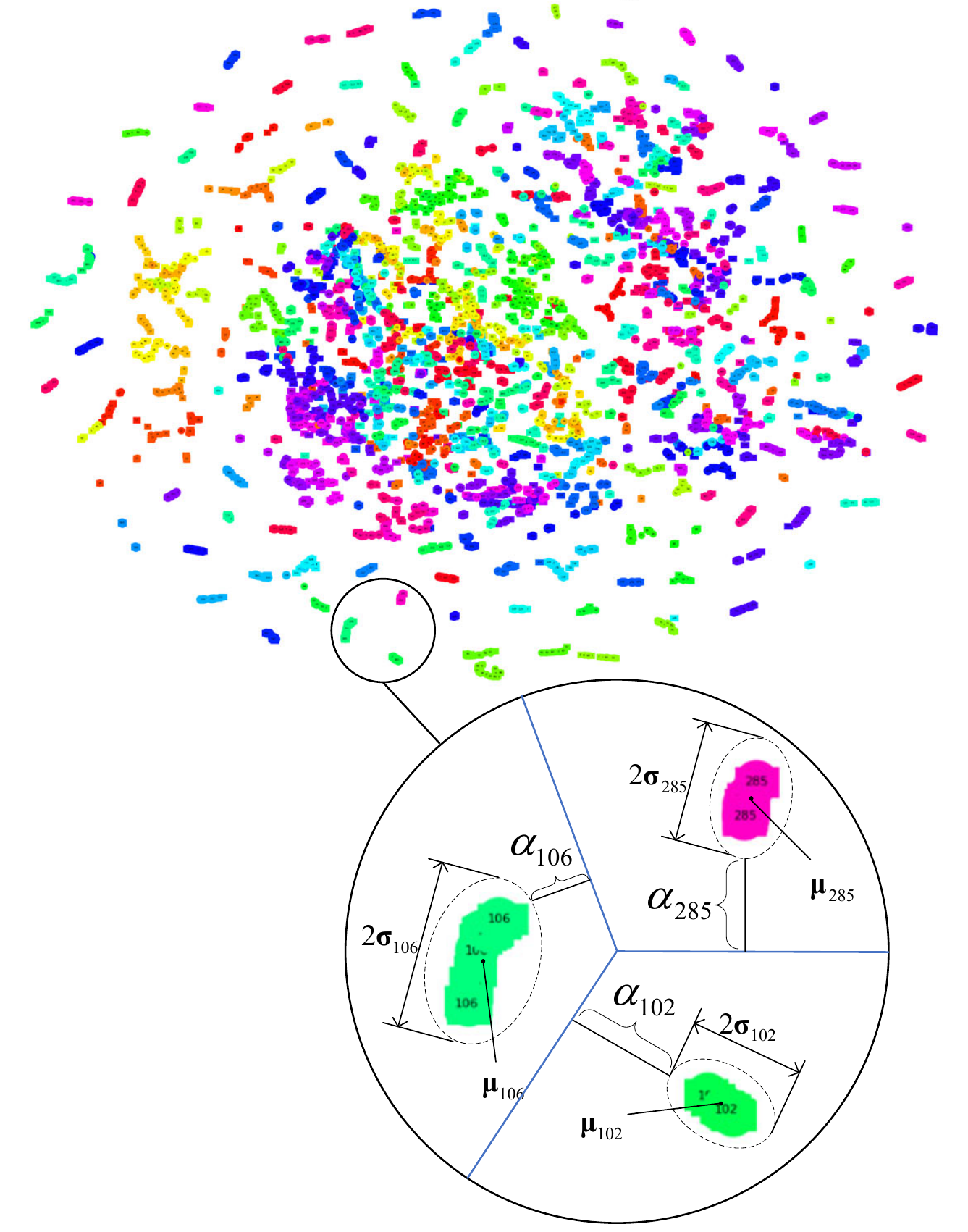}
  \caption{Visualization of embeddings from the test set of SYSU-MM01 dataset. The circles and squares donate the embeddings from IR and RGB images, respectively. The numbers in the circles and squares are the IDs of embeddings. Different colors represent different IDs. The blue line illustrates hyper-plane boundaries. Although the SYSU-MM01 dataset is more challenging than the RegDB dataset, it is shown that the heterogeneous embeddings corresponding to same label are superposed. Meanwhile, the inter-class margin is much larger than the intra-class distance cluster by using our DG-VAE method.}
\end{figure*}

\end{document}